\documentclass[10pt,journal,cspaper,compsoc]{IEEEtran}

\newif\ifMakeReviewDraft
\MakeReviewDraftfalse

\newif\ifUseColorLinks
\UseColorLinkstrue

\newif\ifDisplayMyComments

\DisplayMyCommentstrue
\ifCLASSINFOpdf

\else

\fi

\ifCLASSOPTIONcompsoc
\usepackage[nocompress]{cite}
\else
\usepackage{cite}
\fi

\usepackage{graphicx}
\graphicspath{{./}}
\usepackage{epstopdf}
\usepackage{multirow}
\usepackage{algorithm}
\usepackage{algorithmic}
\renewcommand{\algorithmicrequire}{\textbf{Input:}}
\renewcommand{\algorithmicensure}{\textbf{Output:}}
\renewcommand{\algorithmiccomment}[1]{\textcolor{gray}{\emph{// #1}}} 					
\usepackage{amsmath}
\usepackage{amssymb}
\usepackage{amsthm}
\newtheorem{thm}{Theorem}

\newtheorem{lem}[thm]{Lemma}

\newtheorem{definition}{Definition}

\ifMakeReviewDraft

\usepackage{showkeys}

\fi

\usepackage{booktabs}
\usepackage{subfigure} 
\usepackage{placeins}
\usepackage{verbatim}					
\usepackage{acronym}
\usepackage{color}
\usepackage{url}
\usepackage{hyperref}
\hypersetup{
	pdftitle={},    
	pdfsubject={},	
	pdfauthor={},		
	pdfkeywords={}	
	pdfcreator={LaTeX 2e},	
	pdfproducer={},	
	plainpages=false,			%
	unicode=false,				
	pdftoolbar=true,			
	pdfmenubar=true,			
	pdffitwindow=false,		
	pdfstartview={FitH},	
	pdfnewwindow=true,		
	colorlinks=true				
}
\ifUseColorLinks

\hypersetup{
	linkcolor=blue,			
	citecolor=green,		
	filecolor=magenta,	
	urlcolor=cyan				
}

\else

\hypersetup{
	linkcolor=black,		
	citecolor=black,		
	filecolor=black,		
	urlcolor=black			
}

\fi

\newcommand{\sgn}{\mathop{\mathrm{sgn}}}
\newcommand{\hinge}[1]{\left[ #1 \right]_{+}}

\DeclareMathOperator{\Diagtmp}{diag}
\newcommand{\diag}[1]{\Diagtmp \left( #1 \right) }

\DeclareMathOperator{\Etmp}{\mathbb{E}}

\newcommand{\Es}[2]{\Etmp_{#1} \left\{ #2 \right\} }

\DeclareMathOperator{\Tracetmp}{trace}
\newcommand{\trace}[1]{\Tracetmp \left\{ #1 \right\} }

\definecolor{gray}{rgb}{0.5,0.5,0.5}	
\definecolor{red}{rgb}{1.0,0.0,0.0}	  
\definecolor{green}{rgb}{0.0,1.0,0.0}	  
\definecolor{blue}{rgb}{0.0,0.0,1.0}	  

\newcommand{\ie}{\textit{i.e.}}
\newcommand{\eg}{\textit{e.g.}}

\newtheorem{proposition}{Proposition}

\newacro{MKL}{Multiple Kernel Learning}
\newacro{SVM}{Support Vector Machine}
\newacro{RKHS}{Reproducing Kernel Hilbert Space}
\newacro{MM}{Majorization-Minimization}
\newacro{*SHL}{*Supervised Hash Learning}
\newacro{BCD}{Block Coordinate Descent}
\newacro{MC*SHL}{Multiple Codewords *Supervised Hash Learning}
\newacro{PSD}{Proximal Subgradient Descent}
\newacro{AUC}{Area Under Curve}

\newcommand{\eref}[1]{Eq.~(\ref{#1})}
\newcommand{\eeref}[1]{(\ref{#1})}
\newcommand{\fref}[1]{Fig.~\ref{#1}}
\newcommand{\tref}[1]{Table~\ref{#1}}
\newcommand{\sref}[1]{Sec.~\ref{#1}}
\newcommand{\appref}[1]{Appendix~\ref{#1}}
\newcommand{\aref}[1]{Algorithm~\ref{#1}}
\newcommand{\pref}[1]{Prob.~(\ref{#1})}
\newcommand{\propref}[1]{Prop.~\ref{#1}}
\newcommand{\thmref}[1]{Theorem~\ref{#1}}
\newcommand{\lemmaref}[1]{Lemma~\ref{#1}}

\newcommand{\defref}[1]{Definition~\ref{#1}}
\ifDisplayMyComments

\newcommand{\mycomment}[1] { \noindent {\small \textcolor{blue}{\textbf{Me:} \emph{#1} } } }
\newcommand{\gcacomment}[1] { \noindent {\small \textcolor{red}{\textbf{GCA:} \emph{#1} } } }

\else

\newcommand{\mycomment}[1] {}
\newcommand{\gcacomment}[1] {}

\fi

\begin{document}

%
\title{Learning Hash Function through Codewords}
%
%
%
%

\author{Yinjie~Huang,~\IEEEmembership{Student~Member,~IEEE,}
        Michael~Georgiopoulos,~\IEEEmembership{Member,~IEEE,}
        and~Georgios~C.~Anagnostopoulos,~\IEEEmembership{Member,~IEEE}
\IEEEcompsocitemizethanks{\IEEEcompsocthanksitem Y. Huang and M. Georgiopoulos are with the Department
of Electrical Engineering \& Computer Science, University of Central Florida, Orlando,
FL, 32826.\protect\\
E-mail: yhuang@eecs.ucf.edu, michaelg@ucf.edu
\IEEEcompsocthanksitem G. C. Anagnostopoulos is with the Department of Electrical \& Computer Engineering, Florida Institute of Technology, Melbourne, FL 32901.\protect\\
E-mail: georgio@fit.edu}
\thanks{}}

%
%

\markboth{IEEE TRANSACTIONS ON PATTERN ANALYSIS AND MACHINE INTELLEGENCE,~VOL.~XX, NO.~XX, XXX~2015}%
{Shell \MakeLowercase{\textit{et al.}}: Bare Advanced Demo of IEEEtran.cls for Journals}
%




\IEEEtitleabstractindextext{%
\begin{abstract}
In this paper, we propose a novel hash learning approach that has the following main distinguishing features, when compared to past frameworks. First, the codewords are utilized in the Hamming space as ancillary techniques to accomplish its hash learning task. These codewords, which are inferred from the data, attempt to capture grouping aspects of the data's hash codes. Furthermore, the proposed framework is capable of addressing supervised, unsupervised and, even, semi-supervised hash learning scenarios. Additionally, the framework adopts a regularization term over the codewords, which automatically chooses the codewords for the problem. To efficiently solve the problem, one Block Coordinate Descent algorithm is showcased in the paper. We also show that one step of the algorithms can be casted into several Support Vector Machine problems which enables our algorithms to utilize efficient software package. For the regularization term, a closed form solution of the proximal operator is provided in the paper. A series of comparative experiments focused on content-based image retrieval highlights its performance advantages.
\end{abstract}

}

\maketitle

\IEEEdisplaynontitleabstractindextext

%
\IEEEpeerreviewmaketitle

\ifCLASSOPTIONcompsoc

\acresetall

\IEEEraisesectionheading{\section{Introduction}\label{sec:introduction}}
\else
\section{Introduction}
\label{sec:introduction}
\fi

\IEEEPARstart{W}{ith} the eruptive growth of Internet data including images, music, documents and videos, content-based image retrieval (CBIR) has drawn lots of attention over the past few years \cite{Datta2008}. Given a query sample from a user, a typical CBIR system retrieves samples stored in a database that are most similar to the query sample. The similarity is evaluated in terms of a pre-specified distance metric and the retrieved samples are the nearest neighbors of the query sample w.r.t. this metric. However, in some practical settings, exhaustively comparing the query sample with every sample in the database may be computationally expensive. Furthermore, most CBIR frameworks may be obstructed by the sheer size of each sample; for instance, visual descriptors of an image or a video may contain thousands of features. Additionally, storage of these high-dimensional data also presents a challenge. 

Substantial effort has been invested in designing hash functions transforming the original data into compact binary codes to reap the benefits of a potentially fast similarity search. For example, when compact binary codes in Hamming space used, approximate nearest neighbors (ANN) \cite{Torralba2008} search was shown to achieve sub-liner searching time. Storage of the binary code is, obviously, also much more efficient. Furthermore, hash functions are typically designed to preserve certain similarity qualities between the data in the Hamming space. 

Existing popular hashing approaches can be divided into two categories: \textit{data-independent} and \textit{data-dependent}. While the former category designs the hash function based on a non data-driven approach, the latter category, by inferring from data, can be further clustered into supervised, unsupervised and semi-supervised learning tasks.

In this paper, we propose a novel hash function learning approach\footnote{A preliminary version of the work presented here has appeared in \cite{Huang2015}.}, dubbed \ac{*SHL} (* stands for all three learning paradigms), which exhibits the following advantages: first, the method uses a set of Hamming space codewords, that are learned during training, to capture the intrinsic similarities between the data's hash codes, so that same-class data are grouped together. Unlabeled data also contribute to the adjustment of codewords leveraging from the inter-sample dissimilarities of their generated hash codes, as measured by the Hamming distance metric. Additionally, a regularization term is utilized in our framework to move the codewords which represent the same class closer to each other. When some codewords collapse to one single codeword, our framework achieves automatic selection of the codewords. Due to these codeword-specific characteristics, a major advantage offered by out framework is that it can engage supervised, unsupervised and, even, semi-supervised hash learning tasks using a single formulation. Obviously, the latter ability readily allows the framework to perform transductive hash learning. Note that our framework can be viewed as an Error-Correction Codes (ECOC) method. Readers can refer to \cite{Dietterich1995} and \cite{Mohri2012} for more details of ECOC.

In \sref{sec:Formulation}, we provide \ac{*SHL}'s formulation, which is mainly motivated by an attempt to minimize the within-group Hamming distances in the code space between a group's codeword and the hash codes of data that either should be similar (because of similar labels), or are de-facto similar (due to particular state of the hash functions). With regards to the hash functions, \ac{*SHL} adopts a kernel-based approach. A new regularization term over codewords is also introduced for \ac{*SHL} in its formulation. The aforementioned motivation eventually leads to a minimization problem over the codewords as well as over the \ac{RKHS} vectors defining the hash functions. A quite noteworthy aspect of the resulting formulation is that the minimizations over the latter parameters leads to a set of \ac{SVM} problems, according to which each \ac{SVM} generates a single bit of a sample's hash code. In lieu of choosing a fixed, arbitrary kernel function, we use a simple \ac{MKL} approach (\eg\ see \cite{Kloft2011}) to infer a good kernel from the data. 

Next, in \sref{sec:Optimization}, an efficient \ac{MM} algorithm is showcased that can be used to optimize \ac{*SHL}'s framework via the \ac{BCD} approaches. To train \ac{*SHL}, the first block optimization amounts to training a set of \ac{SVM}, which can be efficiently accomplished by using, for example, \texttt{LIBSVM} \cite{Chang2011}. The second block optimization step addresses the \ac{MKL} parameters. The third block involves solving a problem with the non-smooth regularization over codewords, which is optimized by \ac{PSD}. The second and third blocks are computationally fast thanks to closed-form solutions. When confronted with a huge data set, kernel related problem has computational limitation. In this work, a version of our algorithm for big data, which is based on the software \textit{LIBSKYLARK} \cite{Sindhwani2014}, is also presented.

Finally, in \sref{sec:Experiments} we demonstrate the capabilities of \ac{*SHL} on a series of comparative experiments. The section emphasizes on supervised hash learning problems in the context of CBIR. Additionally, we also apply the semi-supervised version of our framework on the foreground/background interactive image segmentation problems. Remarkably, when compared to other hashing methods on supervised learning hash tasks, \ac{*SHL} exhibits the best retrieval accuracy in all the datasets we considered. Some clues to the method's superior performance are provided in \sref{sec:Generalization}.

\section{Related Work}
\label{sec:RelatedWork}
As mentioned in \sref{sec:introduction}, hashing methods can be divided into two categories: \textit{data-independent} and \textit{data-dependent}. The former category designs the hashing approaches without the necessity to infer from the data. For instance, in \cite{Gionis1999}, Locality Sensitive Hashing (LSH) randomly projects and thresholds data into the Hamming space to generate binary codes. Data samples, which are closely located (in terms of Euclidean distances in the data's native space), are likely to have similar binary codes. Additionally, the authors of \cite{Kulis2009} proposed a method for ANN search through using a learned Mahalanobis metric combined with LSH. \cite{Raginsky2009} introduces an encoding scheme based on random projections, in which the expected Hamming distance between two binary codes of the vectors is related to the value of a shift-invariant kernel. 

On the other hand, \textit{data-dependent methods} can, in turn, be grouped into supervised, unsupervised and semi-supervised learning paradigms.

The majority of work in data-dependent hashing approaches has been studied so far following the supervised learning scenario. For example, Semantic Hashing \cite{Salakhutdinov2009} designs the hash function using a Restricted Boltzmann Machine (RBM). Binary Reconstructive Embedding (BRE), proposed in \cite{Kulis2009a}, tries to minimize a cost function measuring the difference between the original metric distances and the reconstructed distances in the Hamming space. In \cite{Norouzi2011}, through learning the hash functions from pair-wise side information, Minimal Loss Hashing (MLH) formulated the hashing problems based on a bound inspired by the theory of structural Support Vector Machines \cite{Yu2009}. \cite{Mu2010} addresses the scenario, where a small portion of sample pairs are manually labeled as similar or dissimilar and proposes the Label-regularized Max-margin Partition algorithm. Moreover, Self-Taught Hashing \cite{Zhang2010} first identifies binary codes for given documents via unsupervised learning; next, classifiers are trained to predict codes for query documents. Additionally, in \cite{Strecha2012}, Fisher Linear Discriminant Analysis (LDA) was employed to embed the original data to a lower dimensional space and hash codes are obtained subsequently via thresholding. Boosting-based Hashing is used in \cite{Shakhnarovich2003} and \cite{Baluja2008}, in which a set of weak hash functions are learned according to the boosting framework. In \cite{Li2013}, the hash functions are learned from triplets of side information; their method is designed to preserve the relative comparison relationship from the triplets and is optimized using column generation. Furthermore, Kernel Supervised Hashing (KSH) \cite{Liu2012} introduces a kernel-based hashing method, which seems to exhibit remarkable experimental results. Their method utilizes the equivalence between optimizing the code inner products and the Hamming distance. \cite{Lin2014} proposes boosted decision trees for achieving non-linearity in hashing, which is fast to train. Their method employs an efficient GraphCut based block search approach. In \cite{Xia2014}, a supervised hash learning method for image retrieval is designed, in which their method automatically learns a good image representation tailored as well as several hash functions. Latent factor hashing, proposed in \cite{Zhang2014}, learns similarity preserving compact binary codes based on a latent factor model. Finally, \cite{Lin2014a} combines structural Support Vector Machines with hashing methods to directly optimize over multivariate performance measure such as \ac{AUC}.

Several approaches have also been proposed for unsupervised hashing: With the assumption of a uniform data distribution, Spectral Hashing (SPH) \cite{Weiss2008} designs the hash functions by utilizing spectral graph analysis. In \cite{Chen2014}, a new regularization is introduced to control the mismatch between the Hamming codes and the low-dimensional data representation. This new regularizer helps the methods better cope with the data sampled from a nonlinear manifold. Anchor Graph Hashing (AGH)\cite{Liu2011} uses a small-size anchor graph to approximate low-rank adjacency matrices that leads to computational savings. Moreover, \cite{Wang2010} proposed a projection learning method for error correction. Also, in \cite{Gong2011}, the authors introduce Iterative Quantization, which tries to learn an orthogonal rotation matrix so that the quantization error of mapping the data to the vertices of the binary hypercube is minimized. \cite{Liu2014}'s idea is to decompose the feature space into a subspace shared by the hash functions. Then they design an objective function combining spectral embedding loss, binary quantization loss and shared subspace contribution. Finally, \cite{Liu2014a} presents an unsupervised hashing model based on graph model. Their method tries to preserve the neighborhood structure of massive data in a discrete code space.

As for semi-supervised hashing, there are a few approaches proposed: Semi-Supervised Hashing (SSH) in \cite{Wang2010} and \cite{Wang2012} minimizes an empirical error using labeled data; in order to avoid over-fitting, the framework also includes an information theoretic regularizer that utilizes both labeled and unlabeled data. Another method, semi-supervised tag hashing \cite{Wang2014}, incorporates tag information into training hash function by exploring the correlation between tags and hash bits. In \cite{Cheng2014}, the authors introduce a hashing method integrating multiple modalities. Besides, semi-supervised information is also incorporated in the framework and a sequential learning scenario is adopted.

Finally, Let us note here that Self-Taught Hashing (STH) \cite{Zhang2010} employs \acp{SVM} to generate hash codes. However, STH differs significantly from \ac{*SHL}; its unsupervised and supervised learning stages are completely decoupled, while our framework uses a single cost function that simultaneously accommodates both of these learning paradigms. Unlike STH, \ac{SVM}s arise naturally from the problem formulation in \ac{*SHL}.

\section{Formulation}
\label{sec:Formulation}

\subsection{*Supervised Hash Learning}

In what follows, $\left[ \cdot \right]$ denotes the Iverson bracket, \ie, $\left[ \text{predicate} \right] = 1$, if the predicate is true, and $\left[ \text{predicate} \right] = 0$, if otherwise. Additionally, vectors and matrices are denoted in boldface. All vectors are considered column vectors and $\cdot^T$ denotes transposition. Also, for any positive integer $K$, we define $\mathbb{N}_K \triangleq \left\{ 1, \ldots, K \right\}$.

Central to hash function learning is the design of functions transforming data to a compact binary codes in a Hamming space to fulfill a given machine learning task. Consider the Hamming space $\mathbb{H}^B \triangleq \left\{-1, 1\right\}^B$, which implies $B$-bit hash codes. \ac{*SHL} addresses multi-class classification tasks with an arbitrary set $\mathcal{X}$ as sample space. It does so by learning a hash function $\mathbf{h}: \mathcal{X} \rightarrow \mathbb{H}^B$ and a set of $C \times S$ labeled codewords $\boldsymbol{\mu}_{c,s}, \ c \in \mathbb{N}_C$ and $s \in \mathbb{N}_S$ (Each class is represented by $S$ codewords), so that the hash code of a labeled sample is mapped close to the codeword corresponding to the sample's class label, where proximity is measured via the Hamming distance. Unlabeled samples are also able to contribute in learning both the hash function and the codewords as it will be demonstrated in the sequel. Finally, a test sample is classified according to the label of the codeword closest to the sample's hash code.  

In \ac{*SHL}, the hash code for a sample $x \in \mathcal{X}$ is eventually computed as $\mathbf{h}(x) \triangleq \sgn \mathbf{f}(x) \in \mathbb{H}^B$, where the signum function is applied component-wise. Furthermore, $\mathbf{f}(x) \triangleq \left[ f_1(x) \ldots f_B(x) \right]^T$, where $f_b(x) \triangleq \left \langle w_b, \phi(x) \right \rangle_{\mathcal{H}_b} + \beta_b$ with $w_b \in \Omega_{w_b} \triangleq \{ w_b \in \mathcal{H}_b: \left\| w_b \right\|_{\mathcal{H}_b} \leq R_b, R_b>0 \}$ and $\beta_b \in \mathbb{R}$ for all $b \in \mathbb{N}_B$. In the previous definition, $\mathcal{H}_b$ is a \ac{RKHS} with inner product $\left \langle \cdot, \cdot \right \rangle_{\mathcal{H}_b}$, induced norm $\left\| w_b \right\|_{\mathcal{H}_b} \triangleq \sqrt{ \left \langle w_b, w_b \right \rangle_{\mathcal{H}_b} }$ for all $w_b \in \mathcal{H}_b$, associated feature mapping $\phi_b: \mathcal{X} \rightarrow \mathcal{H}_b$ and reproducing kernel $k_b: \mathcal{X} \times \mathcal{X} \rightarrow \mathbb{R}$, such that $k_b(x,x') = \left \langle \phi_b(x), \phi_b(x') \right \rangle_{\mathcal{H}_b}$ for all $x,x' \in \mathcal{X}$. Instead of a priori selecting the the kernel functions $k_b$, \ac{MKL} \cite{Kloft2011} is employed to infer the feature mapping for each bit from the available data. In specific, it is assumed that each \ac{RKHS} $\mathcal{H}_b$ is formed as the direct sum of $M$ common, pre-specified \acp{RKHS} $\mathcal{H}_m$, \ie, $\mathcal{H}_b = \bigoplus_{m} \sqrt{\theta_{b,m}} \mathcal{H}_m$, where $\boldsymbol{\theta}_b \triangleq \left[ \theta_{b,1} \ldots \theta_{b,M} \right]^T \in \Omega_{\theta} \triangleq \left\{ \boldsymbol{\theta} \in \mathbb{R}^M: \boldsymbol{\theta} \succeq \mathbf{0}, \left\| \boldsymbol{\theta} \right\|_p \leq 1, p \geq 1 \right\}$, $\succeq$ denotes the component-wise $\geq$ relation, $\left\| \cdot \right\|_p$ is the usual $l_p$ norm in $\mathbb{R}^M$ and $m$ ranges over $\mathbb{N}_M$. Note that, if each preselected \ac{RKHS} $\mathcal{H}_m$ has associated kernel function $k_m$, then it holds that $k_b(x,x') = \sum_{m} \theta_{b,m} k_m(x,x')$ for all $x,x' \in \mathcal{X}$.  

Now, assume a training set of size $N$ consisting of labeled and unlabeled samples and let $\mathcal{N}_L$ and $\mathcal{N}_U$ be the index sets for these two subsets respectively. Let also $l_n$ for $n \in \mathcal{N}_L$ be the class label of the $n^{th}$ labeled sample. By adjusting its parameters, which are collectively denoted as $\boldsymbol{\omega}$, \ac{*SHL} attempts to reduce the distortion measure

\begin{align}
\label{eq:1}
E(\boldsymbol{\omega}) \triangleq & \sum_{n \in \mathcal{N}_L} \min_{s} d\left( \mathbf{h}(x_n), \boldsymbol{\mu}_{l_n, s} \right) \nonumber \\
& + \sum_{n \in \mathcal{N}_U} \min_{c, s} d\left( \mathbf{h}(x_n), \boldsymbol{\mu}_{c, s} \right)
\end{align}  

\noindent
where $d$ is the Hamming distance defined as $d(\mathbf{h}, \mathbf{h}') \triangleq \sum_{b} \left[ h_b \neq h'_b \right]$. Note that for each sample, one best codeword of each class will be selected to represent it. However, the distortion $E$ is difficult to directly minimize. As it will be illustrated further below, an upper bound $\bar{E}$ of $E$ will be optimized instead.

In particular, for a hash code produced by \ac{*SHL}, it holds that $d\left( \mathbf{h}(x), \boldsymbol{\mu} \right) = \sum_b \left[ \mu^b f_b(x) < 0 \right]$. If one defines $\bar{d}\left( \mathbf{f}, \boldsymbol{\mu} \right) \triangleq \sum_b \hinge{1 - \mu^b f_b}$, where $\hinge{u} \triangleq \max \left\{ 0, u \right\}$ is the hinge function, then $d\left( \sgn \mathbf{f}, \boldsymbol{\mu} \right) \leq \bar{d}\left( \mathbf{f}, \boldsymbol{\mu} \right)$ holds for every $\mathbf{f} \in \mathbb{R}^B$ and any $\boldsymbol{\mu} \in \mathbb{H}^B$. Based on this latter fact, it holds that

\begin{align}
	\label{eq:2}
	E(\boldsymbol{\omega}) & \leq \bar{E}(\boldsymbol{\omega}) \triangleq \sum_{c} \sum_{s} \sum_{n} \gamma_{c,n,s} \bar{d}\left( \mathbf{f}(x_n), \boldsymbol{\mu}_{c,s} \right)
\end{align}

\noindent
where

\begin{align}
	\label{eq:3}
	\gamma_{c,n,s} & \triangleq 
			\begin{cases}
				\left[ c = l_n \right] \left[ s = \arg \min_{s'} \bar{d}\left( \mathbf{f}(x_n), \boldsymbol{\mu}_{l_n, s'} \right) \right] \\
				 \qquad \qquad \qquad \qquad \qquad \qquad \qquad n \in \mathcal{N}_L \\
				 \left[ (c,s) = \arg \min_{c', s'} \bar{d}\left( \mathbf{f}(x_n), \boldsymbol{\mu}_{c', s'} \right)  \right]  \\
				 \qquad \qquad \qquad \qquad \qquad \qquad \qquad n \in \mathcal{N}_U
			\end{cases}
\end{align}

\noindent
It turns out that $\bar{E}$, which constitutes the model's loss function, can be efficiently minimized by a three-step algorithm, which delineated in the next section.

\section{Learning Algorithm}
\label{sec:Optimization}


\subsection{Algorithm for *SHL}
The next proposition allows us to minimize $\bar{E}$ as defined in \eref{eq:2} via a \ac{MM} approach \cite{Hunter2004} and \cite{Hunter2000}. 

\begin{proposition}
\label{prop:1}
For any \ac{*SHL} parameter values $\boldsymbol{\omega}$ and $\boldsymbol{\omega}'$, it holds that

\begin{align}
	\label{eq:4}
	\bar{E}(\boldsymbol{\omega}) \leq \bar{E}(\boldsymbol{\omega} | \boldsymbol{\omega}')  \triangleq \sum_{c} \sum_{s} \sum_{n} \gamma'_{c,n,s} \bar{d}\left( \mathbf{f}(x_n), \boldsymbol{\mu}_{c,s} \right)
\end{align}

\noindent where the primed quantities are evaluated on $\boldsymbol{\omega}'$ and

\begin{align}
	\label{eq:5}
	\gamma'_{c,n,s} & \triangleq 
				\begin{cases}
					\left[ c = l_n \right] \left[ s = \arg \min_{s'} \bar{d}\left( \mathbf{f}'(x_n), \boldsymbol{\mu}'_{l_n, s'} \right) \right] \\
					 \qquad \qquad \qquad \qquad \qquad \qquad \qquad n \in \mathcal{N}_L \\
					 \left[ (c,s) = \arg \min_{c', s'} \bar{d}\left( \mathbf{f}'(x_n), \boldsymbol{\mu}'_{c', s'} \right)  \right]  \\
					 \qquad \qquad \qquad \qquad \qquad \qquad \qquad n \in \mathcal{N}_U
				\end{cases}
\end{align}

\noindent
Additionally, it holds that $\bar{E}(\boldsymbol{\omega} | \boldsymbol{\omega}) = \bar{E}(\boldsymbol{\omega})$ for any $\boldsymbol{\omega}$. In summa, $\bar{E}(\cdot | \cdot)$ majorizes $\bar{E}(\cdot)$.

\end{proposition} 

Its proof is relative straightforward and is based on the fact that for any value of $\gamma'_{c,n,s} \in \left\{0, 1\right\}$ other than $\gamma_{c,n,s}$ as defined in \eref{eq:3}, the value of $\bar{E}(\boldsymbol{\omega} | \boldsymbol{\omega}')$ can never be less than $\bar{E}(\boldsymbol{\omega} | \boldsymbol{\omega}) = \bar{E}(\boldsymbol{\omega})$.

The last proposition gives rise to a \ac{MM} approach, where $\boldsymbol{\omega}'$ are the current estimates of the model's parameter values and $\bar{E}(\boldsymbol{\omega} | \boldsymbol{\omega}')$ is minimized with respect to $\boldsymbol{\omega}$ to yield improved estimates $\boldsymbol{\omega}^*$, such that $\bar{E}(\boldsymbol{\omega}^*) \leq \bar{E}(\boldsymbol{\omega}')$. This minimization can be achieved via a \ac{BCD}, as is argued based on the next proposition.

\begin{proposition}
\label{prop:2}

Minimizing $\bar{E}(\cdot | \boldsymbol{\omega}')$ with respect to the Hilbert space vectors, the offsets $\beta_p$ and the \ac{MKL} weights $\boldsymbol{\theta}_b$, while regarding the codeword parameters as constant, one obtains the following $B$ independent, equivalent problems:

\begin{align}
	\label{eq:6} 
	\underset{\underset{\beta_b \in \mathbb{R}, \boldsymbol{\theta}_b \in \Omega_{\theta}, \mu^b_{c,s} \in \mathbb{H}}{w_{b,m} \in \mathcal{H}_m, m \in \mathbb{N}_M}}{\inf}
		& \lambda_1 \sum_c \sum_s \sum_n \gamma'_{c,n,s} \hinge{ 1 - \mu^b_{c,s} f_b(x_n)  } \nonumber \\  
		& +  \frac{1}{2} \sum_m \frac{ \left\| w_{b,m} \right\|^2_{\mathcal{H}_m} }{\theta_{b,m}} \nonumber \\ 
		& + \lambda_2 \sum_c \sum_{i,j \in S} \left \| \boldsymbol{\mu}_{c,i} - \boldsymbol{\mu}_{c,j} \right \|_2 \ \ \ b \in \mathbb{N}_B
\end{align}

\noindent
where $f_b(x) = \sum_m \left \langle w_{b,m} , \phi_m(x)  \right \rangle_{\mathcal{H}_m} + \beta_b$ and $\lambda_1 > 0$ is a regularization constant.

\end{proposition} 

The proof of this proposition hinges on replacing the (independent) constraints of the Hilbert space vectors with equivalent regularization terms and, finally, performing the substitution $w_{b,m} \gets \sqrt{\theta_{b,m}} w_{b,m}$ as typically done in such \ac{MKL} formulations (\eg\ see \cite{Kloft2011}). The third term in \pref{eq:6} pushes codewords representing the same class closer to each other. With an appropriate value of $\lambda_2$, this regularization helps \ac{*SHL} automatically select the codewords.

Note that \pref{eq:6} is jointly convex with respect to all variables under consideration and, under closer scrutiny, one may recognize it as a binary \ac{MKL} \ac{SVM} training problem, which will become more apparent shortly.  

\textbf{First block minimization:} By considering $w_{b,m}$ and $\beta_b$ for each $b$ as a single block, instead of directly minimizing \pref{eq:6}, one can instead maximize the following problem:

\begin{proposition}
\label{prop:2.5}

The dual form of \pref{eq:6} takes the form of

\begin{align}
\label{eq:7}
\underset{\boldsymbol{\alpha}_b \in \Omega_{a_b}}{\sup} & \ \ \boldsymbol{\alpha}_b^T	\boldsymbol{1}_{NCS} - \frac{1}{2} \boldsymbol{\alpha}_b^T \mathbf{D}_b [(\boldsymbol{1}_{CS} \boldsymbol{1}_{CS}^T) \otimes \mathbf{K}_b] \mathbf{D}_b \boldsymbol{\alpha}_b \ \ \ b \in \mathbb{N}_B
\end{align}

\noindent
where $\mathbf{1}_K$ stands for the all ones vector of $K$ elements ($K \in \mathbb{N}$), $\boldsymbol{\mu}_b \triangleq \left[ \mu_{1,b} \ldots \mu_{C,b} \right]^T$, $\mathbf{D}_b \triangleq \diag{\boldsymbol{\mu}_b \otimes \mathbf{1}_N}$, $\mathbf{K}_b \triangleq \sum_m \theta_{b,m} \mathbf{K}_m$, where $\mathbf{K}_m$ is the data's $m^{th}$ kernel matrix, $\Omega_{a_b} \triangleq \left\{ \boldsymbol{\alpha} \in \mathbb{R}^{NC}: \boldsymbol{\alpha}_b^T (\boldsymbol{\mu}_b \otimes \boldsymbol{1}_N) = 0, \boldsymbol{0} \preceq \boldsymbol{\alpha}_b \preceq \lambda_1 \boldsymbol{\gamma}'  \right\}$ and $\boldsymbol{\gamma}' \triangleq \left[ \gamma'_{1,1,1}, \ldots, \gamma'_{1,N,1}, \gamma'_{1, N, 2}, \ldots, \gamma'_{1, N, S}, \gamma'_{2, N, S}, \ldots, \gamma'_{C, N, S} \right]^T$. 
\end{proposition} 

The detailed proof is provided in \appref{proofprop3}. Given that $\gamma'_{c,n,s} \in \left\{ 0, 1 \right\}$, one can easily now recognize that \pref{eq:7} is a \ac{SVM} training problem, which can be conveniently solved using software packages such as \texttt{LIBSVM}. After solving it, obviously one can compute the quantities $\left \langle w_{b,m} , \phi_m(x)  \right \rangle_{\mathcal{H}_m}$, $\beta_b$ and $\left\| w_{b,m} \right\|^2_{\mathcal{H}_m}$, which are required in the next step. 

When dealing with large scale data sets, the sequential solver \textit{LIBSVM} may encounter the memory bottleneck because of the kernel matrix computation. Therefore a parallel software package is necessary for big data problems. \textit{LIBSKYLARK} \cite{Sindhwani2014}, which utilizes random features \cite{Rahimi2008} to approximate kernel matrix and Alternating Direction Method of Multipliers (ADMM) \cite{Parikh2014a} to parallelize the algorithm, proves to be an efficient solver for large scale \ac{SVM} problem. \textit{LIBSKYLARK} achieves impressive acceleration when solving \ac{SVM} compared to \textit{LIBSVM} in \cite{Sindhwani2014}. Experiments over large data sets are also conducted in \sref{sec:Experiments}.

\textbf{Second block minimization:} Having optimized the \ac{SVM} parameters, one can now optimize the cost function of \pref{eq:6} with respect to the \ac{MKL} parameters $\boldsymbol{\theta}_b$ as a single block using the closed-form solution mentioned in Prop. 2 of \cite{Kloft2011} for $p>1$, which is given below

\begin{align}
	\label{eq:8}
	\theta_{b,m} = \frac{\left \| w_{b,m} \right \|^{\frac{2}{p+1}}_{\mathcal{H}_m}}{ \left( \sum_{m'} \left \| w_{b,m'} \right \|^{\frac{2p}{p+1}}_{\mathcal{H}_{m'}}  \right)^{\frac{1}{p}}}, \ \ \ m \in \mathbb{N}_M, b \in \mathbb{N}_B.
\end{align}

\textbf{Third block minimization:} we need to optimize this problem due to the new regularization introduced:

\begin{align}
\label{eq:3_4}
\underset{\mu^b_{c,s} \in \mathbb{H}}{\inf} \ \ & \sum_n \sum_c \sum_s \gamma'_{c,n,s} \hinge{1 - \mu^b_{c,s} f_b(x_n)} \nonumber \\ 
& + \lambda_2 \sum_c \sum_{i,j \in S} \left \| \boldsymbol{\mu}_{c,i} - \boldsymbol{\mu}_{c,j} \right \|_2 
\end{align}

\noindent
Here, $c \in \mathbb{N}_C, b \in \mathbb{N}_B, s \in \mathbb{N}_S$. 

First of all, we relax $\boldsymbol{\mu}$ to continuous values, similar to relaxing the hashcode as continuous when computing the hinge loss. \eref{eq:3_4} follows the formulation $l(\boldsymbol{x}) + h(\boldsymbol{x})$, which can be solved by proximal methods \cite{Parikh2014}. Since both the terms (hinge loss and regularization) are convex and non-smooth, we employ \ac{PSD} method in a similar fashion as in \cite{Huang2013}, \cite{Chen2009} and \cite{Rakotomamonjy2011}. 

The proximal subgradient descent is 

\begin{align}
\label{eq:3_5}
\boldsymbol{x}^{k+1} := \textbf{prox}_{\eta h}(\boldsymbol{x}^k - \eta \partial l(\boldsymbol{x}^k))
\end{align}

\noindent
where $\eta$ is the step length and $\partial l$ is the subgradient of the function. Here the proximal operator $\textbf{prox}$ is defined as:

\begin{align}
\label{eq:3_6}
\textbf{prox}_{\eta h}(\boldsymbol{v}) \triangleq \arg \min _{\boldsymbol{x}} \ \ \left( h(\boldsymbol{x}) + \frac{1}{2 \eta} \left \| \boldsymbol{x} - \boldsymbol{v} \right \|^2_2 \right)
\end{align}

%
%

To obtain proximal operator, one needs to solve \eref{eq:3_6}. In our problem setting, the regularization is the sum of many non smooth $L_2$ norms, whose closed form proximal operator is not obvious to achieve. Based on the conclusion from \cite{Yu2013}, the proximal operator of sums of the functions can be approximated by sums of the proximal operator of the individual function, i.e. $\textbf{prox}_{\sum h} \approx \sum \textbf{prox}_h$. Thus, all we need is the closed form proximal operator for one individual norm in \eref{eq:3_4}, \ie \ $\sum_{i,j \in S} \left \| \boldsymbol{\mu}_{i} - \boldsymbol{\mu}_{j} \right \|_2$. Let us concatenate all codewords as $\boldsymbol{\mu} = \left[ \boldsymbol{\mu}_1^T, \cdot \cdot \cdot, \boldsymbol{\mu}_S^T \right]^T \in \mathbb{R}^{BS}$. Moreover, a vector is defined as $\boldsymbol{o} \triangleq \left[0, \cdot \cdot \cdot, 1, \cdot \cdot \cdot, -1, \cdot \cdot \cdot, 0 \right ] \in \mathbb{R}^S$, where the value for index $i$ is set to $1$ and $-1$ for index $j$. With the definition of a matrix $U \triangleq \boldsymbol{o} \otimes I_B \in \mathbb{R}^{B \times BS}$, the regularization can be reformulated as $h\left( \boldsymbol{\mu} \right) = \left \| U\boldsymbol{\mu}\right \|_2$, whose proximal operator will be given in the following proposition:

\begin{proposition}
\label{prop:3.5}

Given the norm as: $h\left( \boldsymbol{\mu} \right) = \left \| U\boldsymbol{\mu}\right \|_2$. Following the definition in \eref{eq:3_6}, the proximal operator of this norm:

\begin{align}
\label{eq:3_7}
\textbf{prox}_{\eta h}(\boldsymbol{v}) = 
\begin{cases}
 \boldsymbol{\mu}_1 = \boldsymbol{v}_1 \\ 
 \quad \ \ \vdots \\ 
 \boldsymbol{\mu}_i = \alpha_1 \boldsymbol{v}_i + \alpha_2 \boldsymbol{v}_j\\ 
 \quad \ \ \vdots \\ 
 \boldsymbol{\mu}_j = \alpha_2 \boldsymbol{v}_i + \alpha_1 \boldsymbol{v}_j\\ 
 \quad \ \ \vdots \\
 \boldsymbol{\mu}_S = \boldsymbol{v}_S
\end{cases}
\end{align}

\noindent
where $\alpha_1 = 1 - \alpha_2$ and $\alpha_2 = \min \{ \frac{\eta}{\left \| \boldsymbol{v}_i - \boldsymbol{v}_j \right \|_2}, \frac{1}{2} \}$, $\boldsymbol{v} = \left[ \boldsymbol{v}_1^T, \cdot \cdot \cdot, \boldsymbol{v}_S^T \right]^T$, which is the input vector for proximal operators in \eref{eq:3_6}.
\end{proposition} 

The detailed proof of \propref{prop:3.5} is showcased in \appref{proofprop4}.

Note that, if we consider only one codeword for each class, \pref{eq:3_4} can be simplified without the regularization: 

\begin{align}
\label{eq:33}
\underset{\mu^b_{c} \in \mathbb{H}}{\inf} \ \ & \sum_n \sum_c \gamma'_{c,n} \hinge{1 - \mu^b_{c} f_b(x_n)} 
\end{align}

\noindent \pref{eq:33} can be optimized by mere substitution.

On balance, as summarized in \aref{alg2}, for each bit, the algorithm to \ac{*SHL} consists of one \ac{SVM} optimization and one \ac{MKL} update. For the third step, we evaluate the proximal operator for each regularization and compute the summation to do the \ac{PSD} to optimize codewords. Note that accelerated proximal gradient descent \cite{Beck2009} is utilized here. $\gamma'_{c,n,s}$ is then updated according to the current estimate of the parameters. This algorithm, as shown in \aref{alg2}, continues running until reaching the convergence\footnote{\texttt{MATLAB}\textsuperscript{\textregistered} code of \ac{*SHL}'s algorithm will be made publicly accessible, upon this manuscript's acceptance by the journal.}. Based on \texttt{LIBSVM}, which provides $\mathcal{O}(N^3)$ complexity \cite{List2009}, our algorithm offers the complexity $\mathcal{O}(BN^3)$ per iteration , where $B$ is the code length and $N$ is the number of instances.      


\begin{algorithm}[tb]
 \caption{Optimization of \pref{eq:6}}
 \label{alg2}
 \begin{algorithmic}
 \STATE {\bfseries Input:} Bit Length $B$, Training Samples $X$ containing labeled or unlabled data.
 \STATE {\bfseries Output:} $\boldsymbol{\omega}$.
 \STATE 1. Initialize $\boldsymbol{\omega}$.
 \STATE 2. {\bfseries While Not Converged}
 \STATE 3. \quad {\bfseries For each bit} 
 \STATE 4. \quad \quad $\gamma_{g, n, s}' \leftarrow \eref{eq:5}$.
 \STATE 5. \quad \quad Step 1: Update $w_{b,m}$ and $\beta_b$.
 \STATE 6. \quad \quad Step 2: Compute $\left\| w_{b,m} \right\|^2_{\mathcal{H}_m}$.
 \STATE 7. \quad \quad \quad \quad \quad \ Update $\theta_{b, m}$.
 \STATE 8. \quad \quad Step 3: {\bfseries For} $k = 1, 2,\cdot \cdot \cdot$ {\bfseries do}
 \STATE 9. \quad \quad \quad \quad \quad \quad \ $\boldsymbol{z}^k = \boldsymbol{\mu}^k - \eta \partial l(\boldsymbol{\mu}^k)$.
 \STATE 10. \quad \quad \quad \quad \quad \quad  $\boldsymbol{y}^k = \sum \textbf{prox}_{\eta h}(\boldsymbol{z}_k)$.
 \STATE 11. \quad \quad \quad \quad \quad \quad  $\boldsymbol{\mu}^{k+1} = \boldsymbol{y}^k + \frac{k-1}{k+2}(\boldsymbol{y}^k - \boldsymbol{y}^{k-1})$.
 \STATE 12. \quad \quad \quad \quad \quad  {\bfseries End For}
 \STATE 13. \quad {\bfseries End For}
 \STATE 14. {\bfseries End While}
 \STATE 15. Output $\boldsymbol{\omega}$.
 \end{algorithmic}
 \end{algorithm}

\begin{figure*}[htb]
\vskip 0.2in
\begin{center}
\centerline{\includegraphics[width=\textwidth]{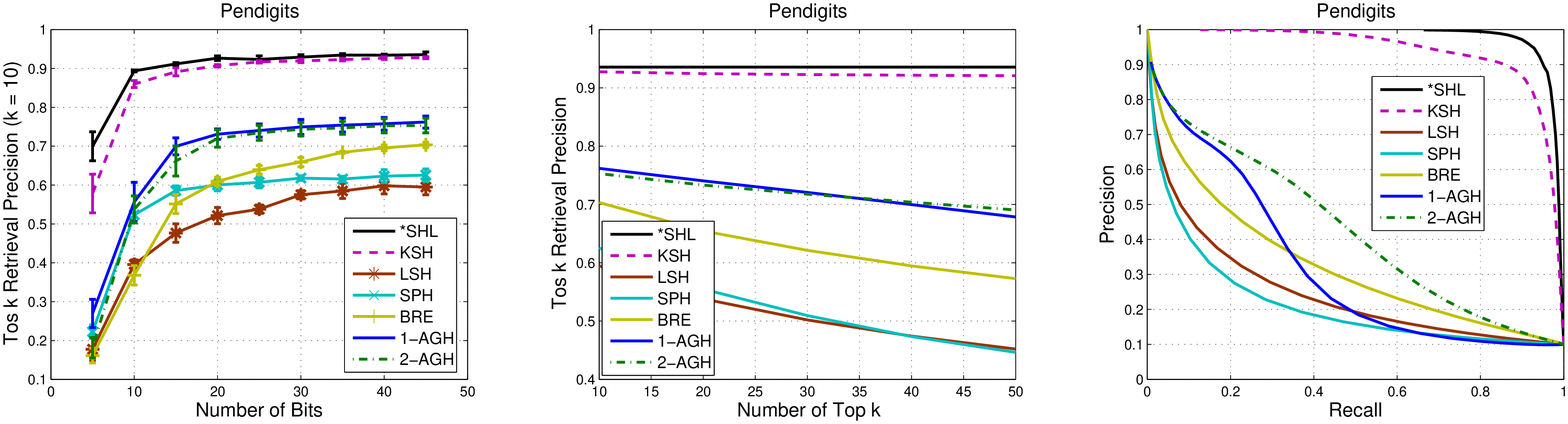}}
\caption{The top $k$ retrieval results and Precision-Recall curve on \textit{Pendigits} dataset over \ac{*SHL} and $6$ other hashing algorithms. (view in color)}
\label{figure2}
\end{center}
\vskip -0.2in
\end{figure*}

\section{Insights to Generalization Performance}
\label{sec:Generalization}

The superior performance of \ac{*SHL} over other state-of-the-art hash function learning approaches featured in the next section can be explained to some extent by noticing that \ac{*SHL} training attempts to minimize the normalized (by $B$) expected Hamming distance of a labeled sample to the correct codeword, which is demonstrated next. We constrain ourselves to the case, where the training set consists only of labeled samples (\ie, $N = \mathcal{N}_L$, $\mathcal{N}_U = 0$) and, for reasons of convenience, to our single codeword \ac{*SHL}. The definitions of the \ac{MKL} hypothesis space for \ac{*SHL} can be found in \sref{sec:Formulation}. Before we provide the generalization results, the following two definitions are necessary.

\begin{figure*}[htb]
\vskip 0.2in
\begin{center}
\centerline{\includegraphics[width=\textwidth]{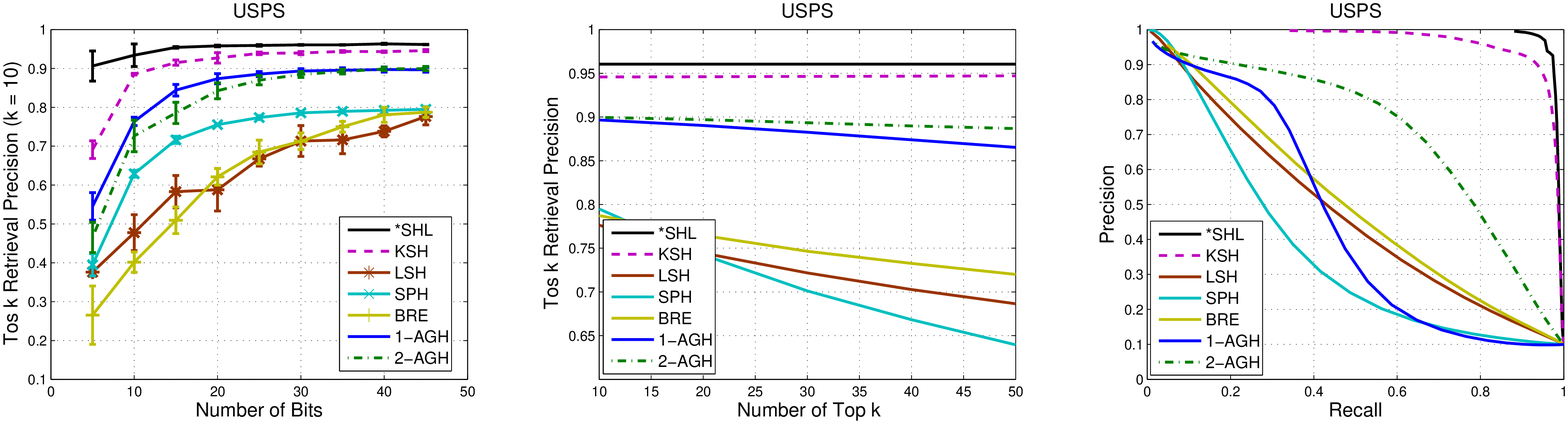}}
\caption{The top $k$ retrieval results and Precision-Recall curve on \textit{USPS} dataset over \ac{*SHL} and $6$ other hashing algorithms. (view in color)}
\label{figure3}
\end{center}
\vskip -0.2in
\end{figure*} 

\begin{figure*}[htb]
\vskip 0.2in
\begin{center}
\centerline{\includegraphics[width=\textwidth]{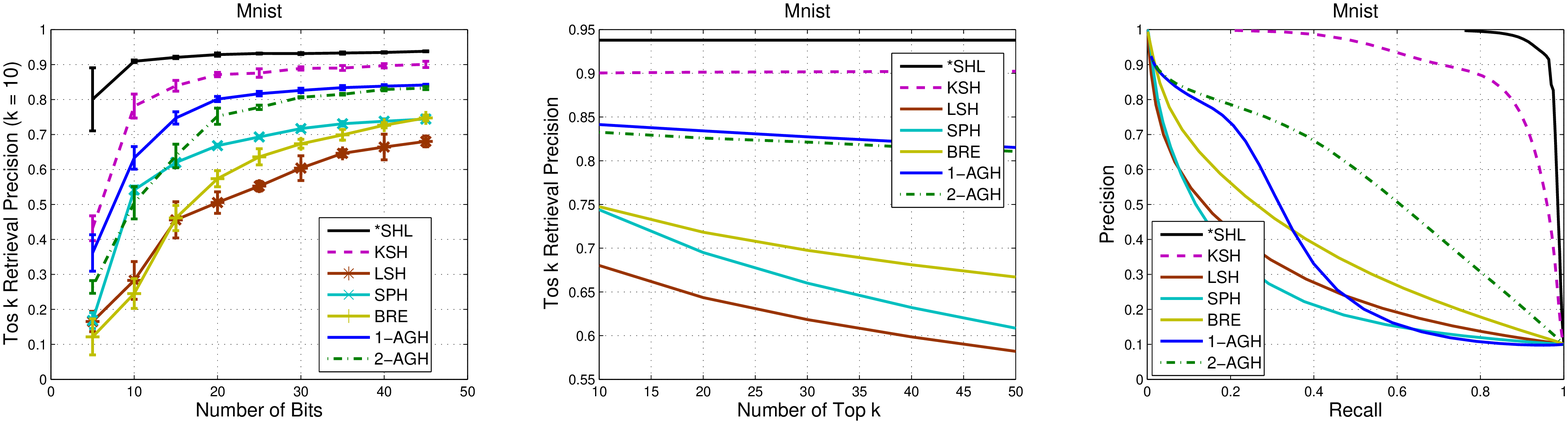}}
\caption{The top $k$ retrieval results and Precision-Recall curve on \textit{Mnist} dataset over \ac{*SHL} and $6$ other hashing algorithms. (view in color)}
\label{figure4}
\end{center}
\vskip -0.2in
\end{figure*} 

\begin{figure*}[htb]
\vskip 0.2in
\begin{center}
\centerline{\includegraphics[width=\textwidth]{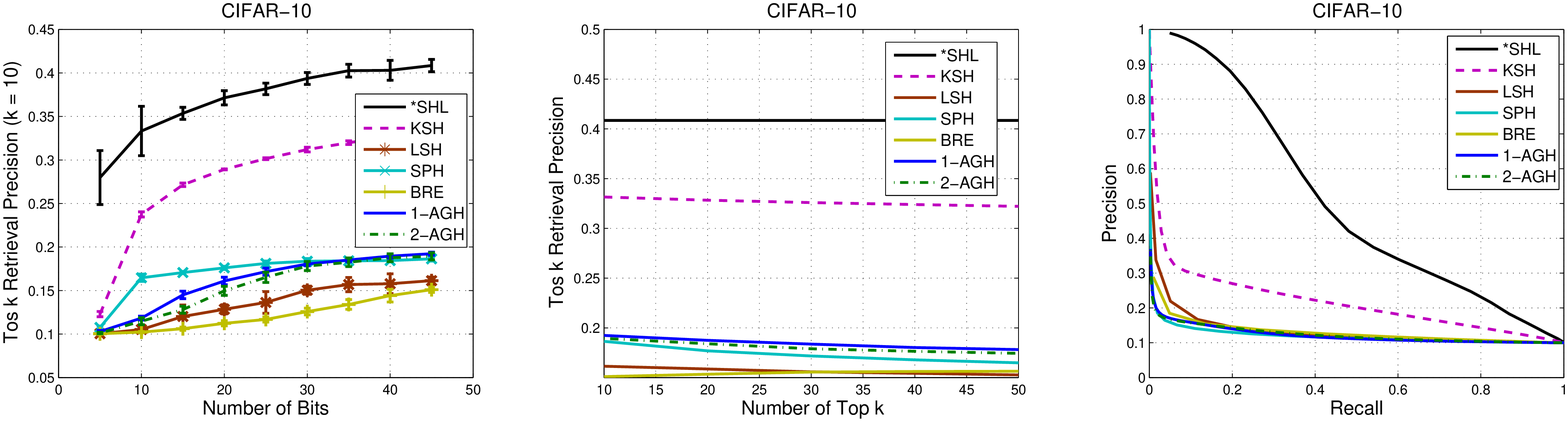}}
\caption{The top $k$ retrieval results and Precision-Recall curve on \textit{CIFAR-10} dataset over \ac{*SHL} and $6$ other hashing algorithms. (view in color)}
\label{figure5}
\end{center}
\vskip -0.2in
\end{figure*} 

\begin{figure*}[htb]
\vskip 0.2in
\begin{center}
\centerline{\includegraphics[width=\textwidth]{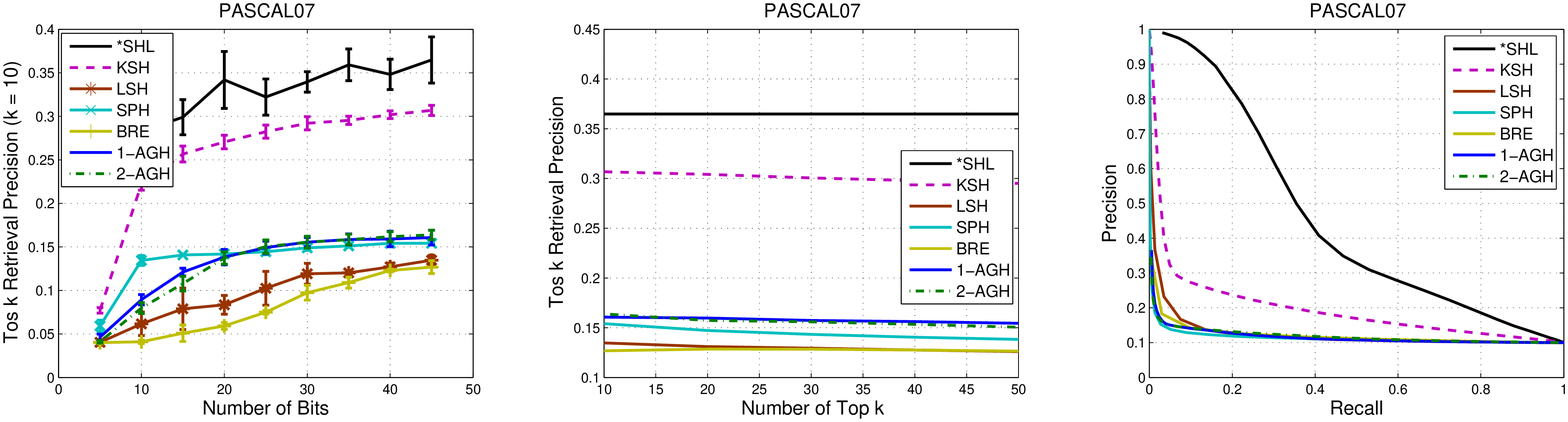}}
\caption{The top $k$ retrieval results and Precision-Recall curve on \textit{PASCAL07} dataset over \ac{*SHL} and $6$ other hashing algorithms. (view in color)}
\label{figure6}
\end{center}
\vskip -0.2in
\end{figure*} 

\begin{definition}
\label{def:def1}
A Random Variable (RV) $\sigma$ that is Bernoulli distributed such that $Pr\{\sigma = \pm 1 \} = \frac{1}{2}$ will be called a Rademacher RV.
\end{definition}

\begin{definition}
\label{def:def2}
Let $\mathcal{G} \triangleq \{ g: \mathcal{Z} \mapsto \mathbb{R} \}$ be a set of functions of an arbitrary domain $\mathcal{Z}$ and $Q = \{ z_n \}^N_{n = 1}$ a set of iid samples from $\mathcal{Z}$ according to a distribution $D$. Then, the Empirical Rademacher Complexity (ERG) of $ \mathcal{G} $ w.r.t $Q$ is defined as: 

\begin{align}
\label{eq:ERC}
\widehat{\Re}_Q(\mathcal{G}) \triangleq \Es{\boldsymbol{\sigma}}{\sup_{g \in \mathcal{G}} \frac{1}{N} \sum_{n = 1}^N \sigma_n o(z_n) \bigg\vert Q }
\end{align}

\noindent
where $\Es{\boldsymbol{\sigma}}{\cdot} \triangleq \Es{\sigma_1}{\Es{\sigma_2}{\cdot \cdot \cdot \Es{\sigma_N}{\cdot} \cdot \cdot \cdot}}$ and $ \{ \sigma_n\}^N_{n = 1}$ are iid Rademacher RV. In the rest of the section, the condition on $Q$ will be omitted for simplicity. Additionally, the Rademacher Complexity (RC) of $\mathcal{G}$ for a sample size $N$ is defined as:

\begin{align}
\label{eq:RC}
\Re_N(\mathcal{G}) \triangleq \Es{Q \sim D^N}{\widehat{\Re}_Q(\mathcal{G})}
\end{align}

\end{definition}

We also need the following two lemmas before showing our final concentration results.

\begin{lem}
\label{lemma:1}
Let $\mathcal{Z}$ be an arbitrary set, $\widetilde{\mathcal{F}} \triangleq \{ \mathbf{f}: z \mapsto \mathbf{f}(z) \in \mathbb{R}^B, \ z \in \mathcal{Z} \}$, $\Psi: \mathbb{R}^B \rightarrow \mathbb{R}$ be L-Lipschitz continuous w.r.t $\left \| \cdot \right \|_1$. Also, define $\Psi \circ \widetilde{\mathcal{F}} \triangleq \{ g: z \mapsto \Psi(\boldsymbol{f}(z))\}$ and $\left \| \widetilde{\mathcal{F}} \right \|_1 \triangleq \{h: z \mapsto \left \| \boldsymbol{f}(z) \right \|_1, \ \boldsymbol{f} \in \widetilde{\mathcal{F}} \}$ then

\begin{align}
\label{eq:4-1}
\widehat{\Re}_Q \left( \Psi \circ \widetilde{\mathcal{F}} \right) \leq L\widehat{\Re}_Q \left( \left \| \widetilde{\mathcal{F}} \right \|_1 \right)
\end{align}

\noindent 
where $Q$ is a set of N samples drawn from $\mathcal{Z}$.
\end{lem}

\begin{lem}
\label{lemma:2}
Let $\mathcal{Z}$ be an arbitrary set. Define: $\widetilde{\mathcal{F}} \triangleq \{ \boldsymbol{f}: z \mapsto \boldsymbol{f}(z) \in \mathbb{R}^B, \ z \in \mathcal{Z} \}$, $\left \| \widetilde{\mathcal{F}}\right \|_1 \triangleq \{ h: z \mapsto \left \| \boldsymbol{f}(z) \right \|_1, \ \boldsymbol{f} \in \widetilde{\mathcal{F}} \}$ and $\boldsymbol{1}^T \widetilde{\mathcal{F}} \triangleq \{ g: z \mapsto \boldsymbol{1}^T \boldsymbol{f}(z), \ \boldsymbol{f} \in \widetilde{\mathcal{F}} \}$. Let's further assume that if $\boldsymbol{f}(z) \triangleq [ f_1(z), ..., f_B(z) ]^T \in \widetilde{\mathcal{F}}$ for $z \in \mathcal{Z}$, then also $[\pm f_1(z), ..., \pm f_B(z)] \in \widetilde{\mathcal{F}}$ for any combination of signs. Then:

\begin{align}
\label{eq:lemma2}
\widehat{\Re}_Q \left(\left \| \widetilde{\mathcal{F}} \right \|_1 \right) \leq \widehat{\Re}_Q \left( \boldsymbol{1}^T \widetilde{\mathcal{F}} \right)
\end{align}

\noindent
where $Q$ is a set of $N$ samples drawn from $\mathcal{Z}$.
\end{lem}

\noindent
The detailed proof is provided in \appref{prooflemma1} for \lemmaref{lemma:1} and in \appref{prooflemma2} for \lemmaref{lemma:2}.

\noindent
To show the main theoretical result of our paper with the help of the previous lemmas, we will consider the sets of functions

\begin{align}
\label{eq:4-2}
\bar{\mathcal{F}} \triangleq & \{ \mathbf{f}: x \mapsto [f_1(x), ..., f_B(x)]^T, \ f_b \in \mathcal{F}, \ b \in \mathbb{N}_B\} \\
\label{eq:4-3}
\mathcal{F} \triangleq & \{ f: x \mapsto \left\langle w, \ \phi_{\boldsymbol{\theta}}(x) \right\rangle_{\mathcal{H}_{\boldsymbol{\theta}}} + \beta, \ \  \beta \in \mathbb{R}, \  \boldsymbol{w} \in \Omega_w(\boldsymbol{\theta}) \nonumber \\ 
&, \ \boldsymbol{\theta} \in \Omega_{\boldsymbol{\theta}} \}
\end{align}

\noindent
where $\Omega_w(\boldsymbol{\theta}) \triangleq \{ w \in \mathcal{H}_{\boldsymbol{\theta}}: \ \left \| w \right \|_{\mathcal{H}_{\boldsymbol{\theta}}} \leq R \}$ for some $R \geq 0$ and $\Omega_{\boldsymbol{\theta}} \triangleq \{ \boldsymbol{\theta} \in \mathbb{R}^M : \ \boldsymbol{\theta} \succeq \boldsymbol{0}, \ \left \| \boldsymbol{\theta} \right \|_p \leq 1 \}$ for some $p \geq 1$.

\noindent
\begin{thm}
\label{theorem1}
Assume reproducing kernels of $\{ \mathcal{H}_m \}^M_{m=1}$ s.t. $\ k_m(x,x') \leq r^2, \ \forall x, x' \in \mathcal{X}$ and a set $Q$ of iid samples $Q = \{ (x_1, l_1), ..., (x_N, l_N) \}$. Then for $\rho > 0$ independent of $Q$, for any $\mathbf{f} \in \bar{\mathcal{F}}$, any $\boldsymbol{\mu} \in \mathcal{M}$, where $\mathcal{M} \triangleq \{ \boldsymbol{\mu}: \ \mathbb{N}_C \rightarrow \mathbb{H}^B \}$ and any $ 0 < \delta < 1$, with probability $1 - \delta$, it holds that:

\begin{align}
\label{eq:4-33}
er\left(\mathbf{f}, \boldsymbol{\mu} \right) \leq \widehat{er} \left ( \mathbf{f}, \boldsymbol{\mu} \right) + \frac{2R}{\rho}\sqrt{rM^{\frac{1}{p'}}\sqrt{\frac{p'}{N^3}}} + \sqrt{\frac{\log \left( \frac{1}{\delta} \right)}{2 N}}
\end{align}

\noindent
where $er \left( \mathbf{f}, \boldsymbol{\mu} \right) \triangleq \frac{1}{B} \mathbb{E} \{ d \left( \sgn \mathbf{f}(x), \boldsymbol{\mu}(l) \right)  \}$, $l \in \mathbb{N}_C$ is the true label of $x \in \mathcal{X}$, $\widehat{er}\left( \mathbf{f}, \boldsymbol{\mu}_l \right) \triangleq \frac{1}{N B} \sum_{n, b} \Phi_{\rho} \left( f_b(x_n)\mu_{l_n, b} \right)$, where $\Phi_{\rho}(u) \triangleq \min\left\{1, \max\left\{0, 1 - \frac{u}{\rho} \right\}  \right\}$ and $p' \triangleq \frac{p}{p-1}$.
\end{thm}

\noindent
The detailed proofs for \thmref{theorem1} can be found in \appref{prooftheorem}.

\section{Experiments}
\label{sec:Experiments}

\subsection{Supervised Hash Learning Results}
\label{compareison}

In this section, we compare \ac{*SHL} to other state-of-the-art hashing algorithms: 

\begin{itemize}
\item Kernel Supervised Learning (KSH)\footnote{\url{http://www.ee.columbia.edu/ln/dvmm/downloads/WeiKSHCode/dlform.htm}} \cite{Liu2012}.
\item Binary Reconstructive Embedding (BRE)\footnote{\url{http://web.cse.ohio-state.edu/~kulis/bre/bre.tar.gz}} \cite{Kulis2009a}.
\item single-layer Anchor Graph Hashing (1-AGH) and its two-layer version (2-AGH)\footnote{\url{http://www.ee.columbia.edu/ln/dvmm/downloads/WeiGraphConstructCode2011/dlform.htm}} \cite{Liu2011}.
\item Spectral Hashing (SPH)\footnote{\url{http://www.cs.huji.ac.il/~yweiss/SpectralHashing/sh.zip}} \cite{Weiss2008}.
\item Locality-Sensitive Hashing (LSH) \cite{Gionis1999}.
\end{itemize}

\noindent Five datasets, which are widely utilized in other hashing papers as benchmarks, were considered: 

\begin{itemize}
\item \textit{Pendigits}: a digit dataset ($10,992$ samples, $256$ features, $10$ classes) of $44$ writers from the \textit{UCI Repository}\footnote{\url{http://archive.ics.uci.edu/ml/}}. In our experiment, we randomly choose $3,000$ for training and the rest for testing.
\item \textit{USPS}: a digit dataset also from the \textit{UCI Repository}, is numeric data from the scanning of handwritten digits from envelops by the U.S. Postal Service. Among the dataset ($9,298$ samples, $256$ features, $10$ classes), $3000$ were used for training and others for testing.
\item \textit{Mnist}\footnote{\url{http://yann.lecun.com/exdb/mnist/}}: a hand written digit dataset which contains $70,000$ samples, $784$ features and $10$ classes. The digits have been size-normalized and centered. In our experiment, $6,000$ for training and $64,000$ for testing.
\item \textit{CIFAR-10}\footnote{\url{http://www.cs.toronto.edu/~kriz/cifar.html}}: a labeled image dataset collected from \textit{80 million tiny images} \footnote{\url{http://groups.csail.mit.edu/vision/TinyImages/}}. The dataset consists of $60,000$ samples, $1,024$ features, $10$ classes. We use $6,000$ for training and the rest for testing.
\item \textit{PASCAL07}\footnote{\url{http://pascallin.ecs.soton.ac.uk/challenges/VOC/}} \cite{Everingham2010}: a dataset contains annotated consumer photographs collected from the flickr photo sharing website. The dataset consists of $6878$ samples, $1,024$ features after down-sampling the images, $10$ classes. Here, $3,000$ for training and others for testing. 
\end{itemize}

For all the algorithms used, average performances over $5$ runs are reported in terms of the following two criteria: (i) retrieval precision of $k$-closest hash codes of training samples; we used $k=\left\{ 10, 15, \ldots, 50 \right\}$. (ii) Precision-Recall (PR) curve, where retrieval precision and recall are computed for hash codes within a Hamming radius of $r \in \mathbb{N}_B$. 

The following \ac{*SHL} settings were used: \ac{SVM}'s parameter $\lambda_1$ was set to $1000$; for \ac{MKL}, $11$ kernels were considered: $1$ normalized linear kernel, $1$ normalized polynomial kernel and $9$ Gaussian kernels. For the polynomial kernel, the bias was set to $1.0$ and its degree was chosen as $2$. For the bandwidth $\sigma$ of the Gaussian kernels the following values were used: $[2^{-7},2^{-5},2^{-3},2^{-1},1,2^1,2^3,2^5,2^7]$. Regarding the \ac{MKL} constraint set, a value of $p = 2$ was chosen. $\lambda_2$ was set to $2000$ for \textit{Pendigits}, \textit{USPS} and \textit{PASCAL07} and $\lambda_2 = 6000$ for the rest of the datasets.


\begin{figure*}[htb]
\vskip 0.2in
\begin{center}
\centerline{\includegraphics[width=\textwidth]{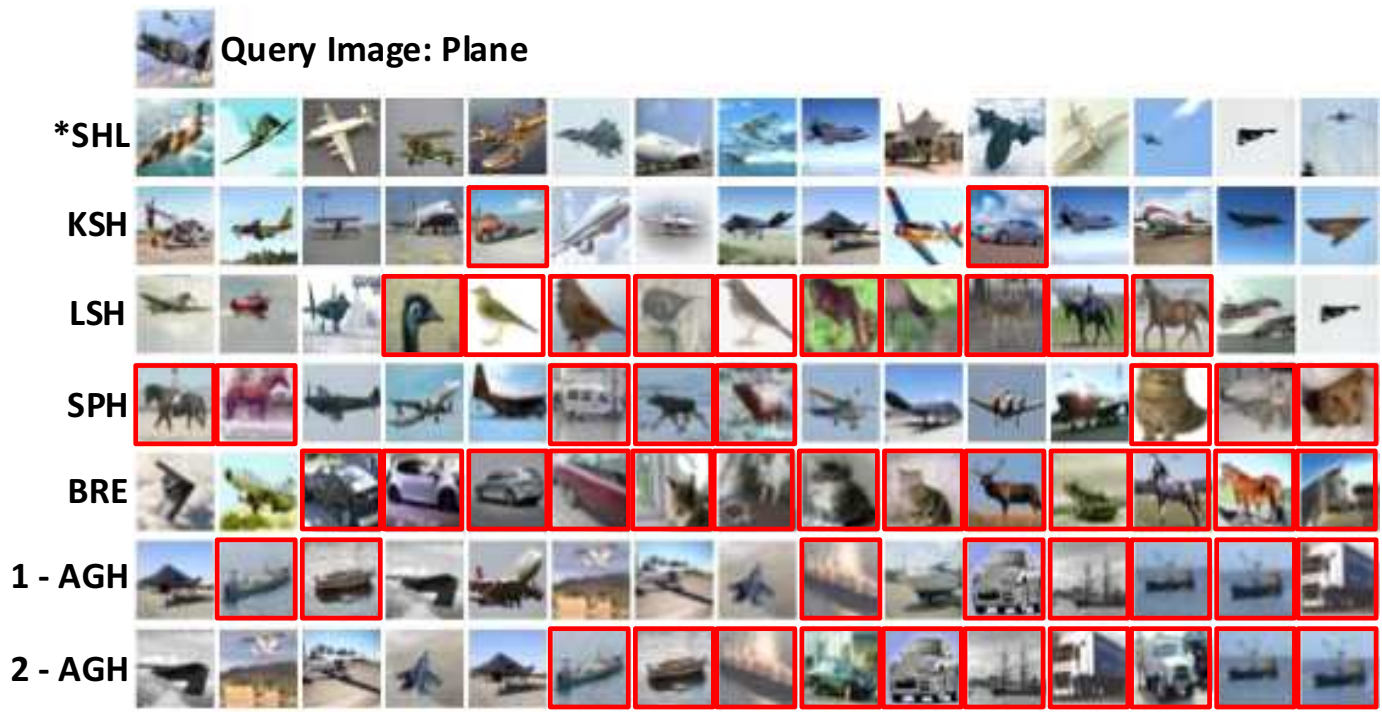}}
\caption{Qualitative results on CIFAR-10. Query image is "Plane". The remaining $15$ images for each row were retrieved using 45-bit binary codes generated by different hashing algorithms. Red box indicates wrong retrieval results.}
\label{figure7}
\end{center}
\vskip -0.2in
\end{figure*} 

For the remaining approaches, namely KSH, SPH, AGH, BRE, parameter values were used according to recommendations found in their respective references. All obtained results are reported in \fref{figure2} through \fref{figure6}.

We clearly observe that \ac{*SHL} performs the best among all the algorithms considered. For all the datasets, \ac{*SHL} achieves the highest top-$10$ retrieval precision. Especially for non-digits datasets (\textit{CIFAR-10}, \textit{PASCAL07}), \ac{*SHL} achieves significantly better results. As for the PR-curve, \ac{*SHL} also obtains the largest areas under the curve. Although impressive results have been reported in \cite{Liu2012} for KSH, in our experiments, \ac{*SHL} outperforms it across all datasets. Moreover, we observe that supervised hash learning algorithms, except BRE, perform better than unsupervised variants. BRE may need a longer bit length to achieve better performance like in \fref{figure2} and \fref{figure4}. Additionally, it is worth mentioning that \ac{*SHL} performs impressively with short bit length across all the datasets.  

AGH also yields good results, compared with other unsupervised hashing algorithms, because it utilizes anchor points as side information to generate hash codes. With the exception of \ac{*SHL} and KSH, the remaining approaches exhibit poor performance for the non-digits datasets we considered (\textit{CIFAR-10} and \textit{PSACAL07}).

\begin{figure*}[htb]
\vskip 0.2in
\begin{center}
\centerline{\includegraphics[width=\textwidth]{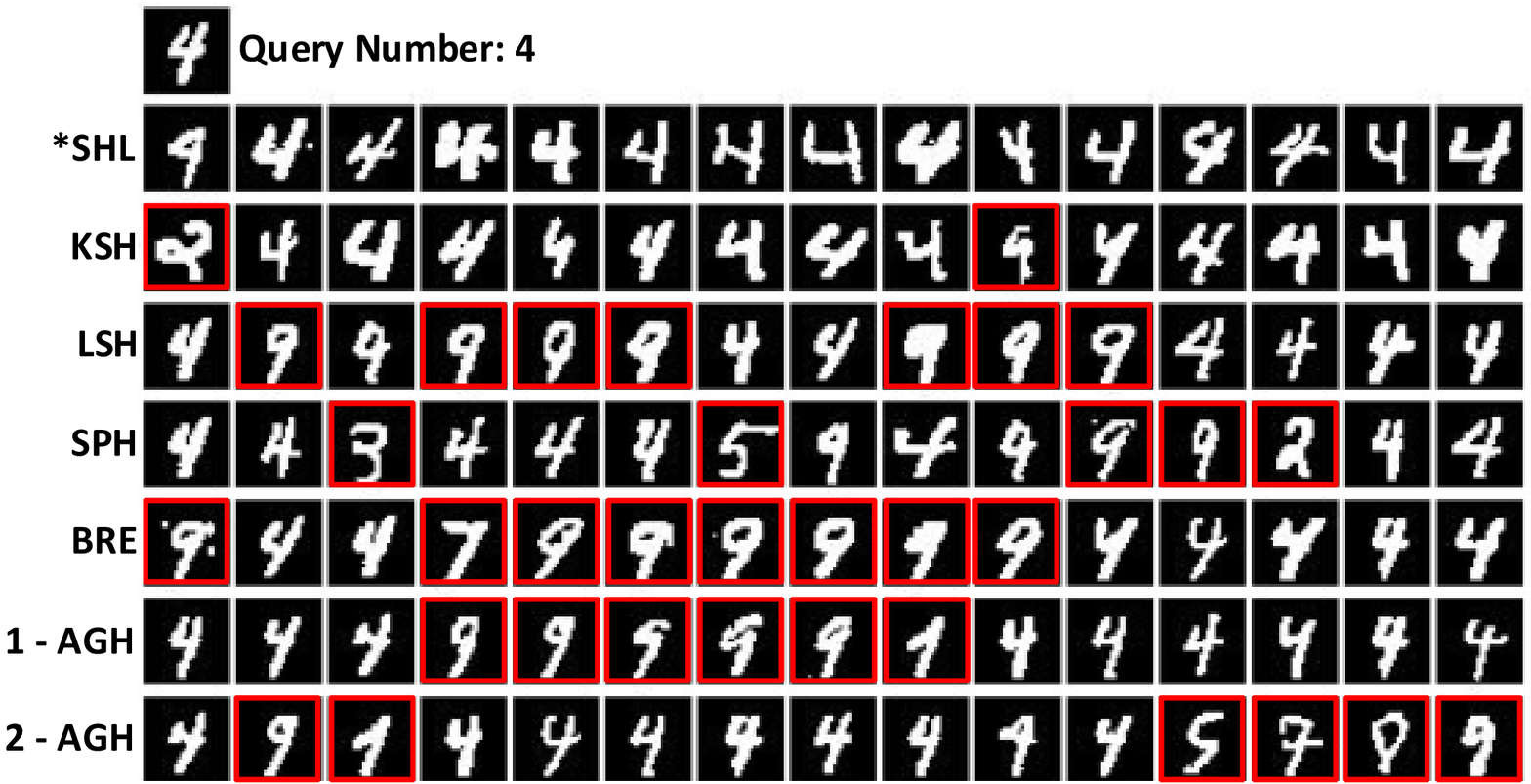}}
\caption{Qualitative results on Mnist. Query image is "4". The remaining $15$ images for each row were retrieved using 45-bit binary codes generated by different hashing algorithms. Red box indicates wrong retrieval results.}
\label{figure8}
\end{center}
\vskip -0.2in
\end{figure*}

When varying the top-$k$ number between $10$ and $50$, once again with the exception of \ac{*SHL} and KSH, the performance of the remaining approaches deteriorated in terms of top-$k$ retrieval precision. KSH performs slightly worse, when $s$ increases, while \ac{*SHL}'s performance remain robust for \textit{CIFAR-10} and \textit{PSACAL07}. It is worth mentioning that the two-layer AGH exhibits better robustness than its single-layer version for datasets involving images of digits. Finally, \fref{figure7} and \fref{figure8} show some qualitative results for the \textit{CIFAR-10} and \textit{Mnist} datasets. In conclusion, it seems that both \ac{*SHL}'s performances are superior at every code length we considered.

\subsection{Transductive Hash Learning Results}
\label{toy}

\begin{figure*}[ht]
\vskip 0.2in
\begin{center}
\centerline{\includegraphics[width=\textwidth]{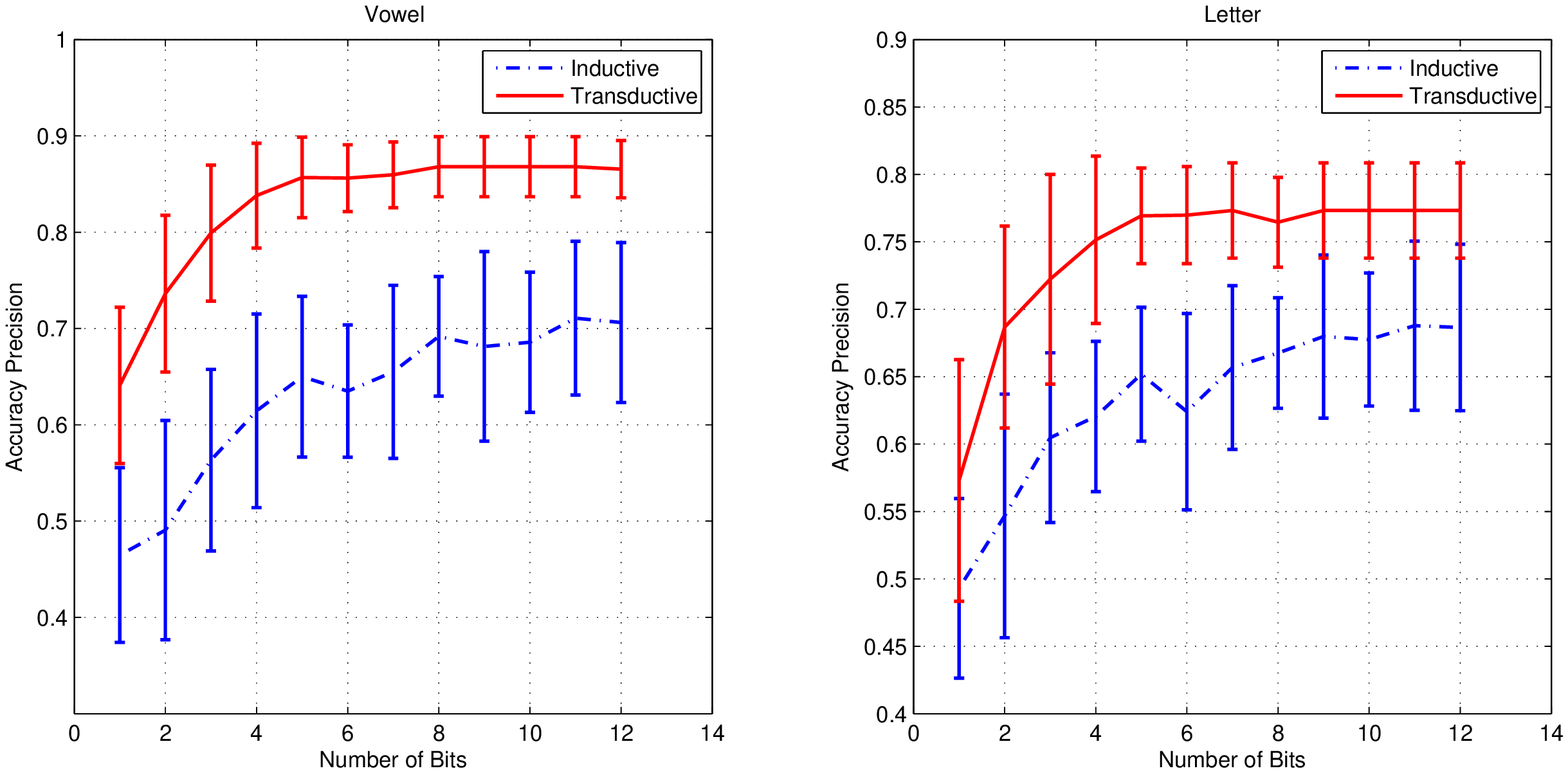}}
\caption{Accuracy results between Inductive and Transductive Learning.}
\label{figure1}
\end{center}
\vskip -0.2in
\end{figure*}

As a proof of concept, in this section, we report a performance comparison of \ac{*SHL}, when used in an inductive versus a transductive \cite{Vapnik1998} mode. Note that, to the best of our knowledge, apart from our method, there are no other hash learning approaches to date that can accommodate transductive hash learning. For illustration purposes, we used the \textit{Vowel} and \textit{Letter} datasets from \textit{UCI Repository}. We randomly chose $330$ training and $220$ test samples for the \textit{Vowel} and $300$ training and $200$ test samples for the \textit{Letter}. Each scenario was run $20$ times and the code length ($B$) varied from $4$ to $15$ bits. The results are shown in \fref{figure1} and reveal the potential merits of the transductive \ac{*SHL} learning mode across a range of code lengths. 

\subsection{Image Segmentation}

\begin{figure*}[htb]
\vskip 0.2in
\begin{center}
\centerline{\includegraphics[scale = 0.85]{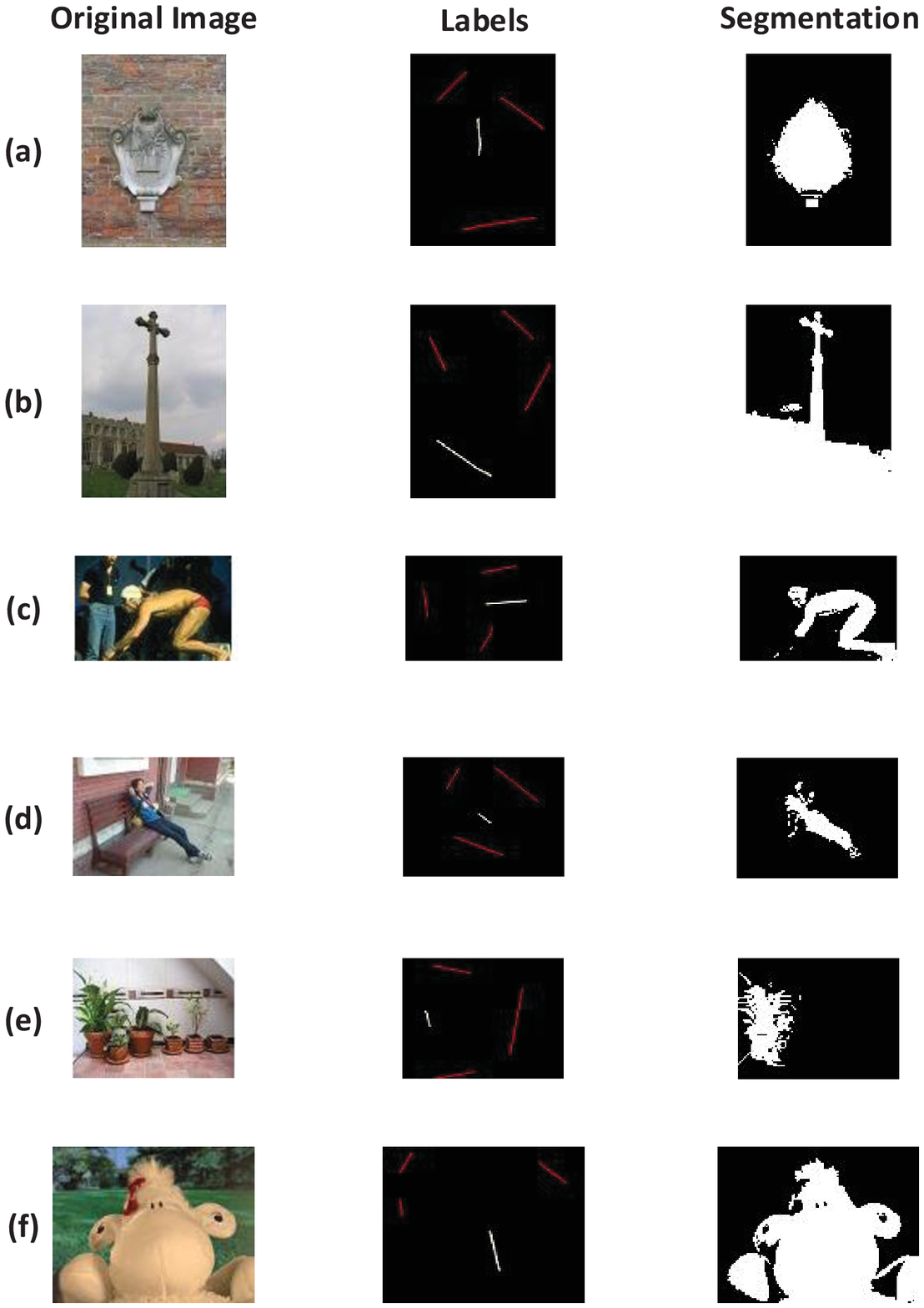}}
\caption{Foreground/Background interactive image segmentation. The left column contains the original images. The middle column includes labeled pixels. The right column shows the results of the segmentati
	on.}
\label{figure9}
\end{center}
\vskip -0.2in
\end{figure*} 

Besides content-based image retrieval, the proposed \ac{*SHL} can also be utilized in the other applications, for example, the foreground/background interactive image segmentation \cite{Blake2004}, where the images are partially labeled as foreground and background by users. In \ac{*SHL}, while foreground and background are represented by two codewords, the rest of the pixels can be labeled in semi-supervised learning scenario. In this section, we show the interactive image segmentation results using \ac{*SHL} on the dataset introduced in \cite{Gulshan2010}. The hash code length is $5$ and the rest of the parameters settings follow the previous section. For each pixel, the RGB values are used as features. The results are shown in \fref{figure9}. We notice that, provided with partially labeled information, \ac{*SHL} successfully segment the foreground object from the background. Especially in (e), although all the flower pots share the same color, \ac{*SHL} only highlight the labeled one and its plant. Additionally, in some images, like (c) and (f), shaded areas fail to be segmented. In these cases, more pixel features may be necessary for better results. 

\subsection{*SHL for Large Data Set}

Large data sets require a huge kernel matrix which can not be fit into the memory of one single machine. Thus, \textit{LIBSKYLARK}, a parallel machine learning software which utilizes kernel approximation and ADMM, replaces \textit{LIBSVM} in \ac{*SHL}. In this section, we created our own clusters in \textit{Amazon Web Service}\footnote{\url{https://aws.amazon.com/}}. One cluster consists of $2$ nodes while the other one contains $10$ nodes. Each node has Xeon E5-2666 CPU and $3.75$ GB memory.

Three data sets are considered here: 

\begin{itemize}
	\item \textit{USPS}: with $256$ features, $7291$ samples are used for training, while $2007$ samples are for testing.
	\item \textit{Mnist}: $60$K are for training and $10$K are for testing. This data set contains $784$ features.
	\item \textit{Mnist1M}: $784$ features. $1$ million samples for training and $100$K for testing.
\end{itemize}

\textit{LIBSKYLARK}\footnote{\url{http://xdata-skylark.github.io/libskylark/}} is downloaded and compiled. We run the experiments using $5$ bits, $25$ bits and $45$ bits of the codeword for the three data sets. \ac{*SHL}'s parameters are set as the previous sections. The results of running time and top-$10$ retrieval accuracies are reported in \tref{table1}:

\begin{table*}
	\setlength{\tabcolsep}{5pt}
	\caption{Top-$10$ retrieval results and running time for \ac{*SHL} for various data sets. Here, running time secs means seconds, mins means minutes and hours means hours.}
	\label{table1}
	\vskip 0.15in
	\begin{center}
			\begin{sc}
				\begin{tabular}{lcccccc}
					\hline
					\noalign{\smallskip}
					Data sets & Nodes & Training & Testing & Bits & Accuracy & Time \\
					\noalign{\smallskip}
					\hline
					\noalign{\smallskip}
					\multirow{3}{*}{\textit{USPS}} & \multirow{3}{*}{$2$} & \multirow{3}{*}{$7291$} & \multirow{3}{*}{$2007$} & $5$	 & $0.916$ & $45.82 \ secs$	\\
												   & & & & $25$ & $0.936$ & $227.17 \ secs$ / $3.78 \ mins$ \\
										    	   & & & & $45$ & $0.941$ & $408.66 \ secs$ / $6.90 \ mins$ \\
					\hline
					\multirow{3}{*}{\textit{Mnist}} & \multirow{3}{*}{$2$} & \multirow{3}{*}{$60K$} & \multirow{3}{*}{$10K$} & $5$ & $0.826$ & $1690.46 \ secs$ / $28.17 \ mins$	\\
											       & & & & $25$ & $0.960$ & $8406.04 \ secs$ / $140.08 \ mins$ / $2.33 \ hours$ \\
												   & & & & $45$ & $0.969$ & $4253.28 \ secs$ / $251.85 \ mins$ / $4.20 \ hours$ \\
					\hline 
					\multirow{3}{*}{\textit{Mnist}}	& \multirow{3}{*}{$10$} & \multirow{3}{*}{$60K$} & \multirow{3}{*}{$10K$} & $5$ & $0.839$ & $487.88 \ secs$ / $8.13 \ mins$ \\
												   & & & & $25$ & $0.962$ & $2361.07 \ secs$ / $39.35 \ mins$ \\
												   & & & & $45$ & $0.964$ & $4253.28 \ secs$ / $70.89 \ mins$ / $1.18 \ hours$ \\
					\hline
					\multirow{3}{*}{\textit{Mnist1M}} & \multirow{3}{*}{$10$} & \multirow{3}{*}{$1M$} & \multirow{3}{*}{$100K$}	& $5$ & $0.824$ & $7918.38 \ secs$ / $131.97 \ mins$ / $2.20 \ hours$ \\ 
									         	   & & & & $25$ & $0.927$ & $28066.37 \ secs$ / $467.77 \ mins$ / $7.80 \ hours$ \\
												   & & & & $45$ & $0.934665$  & $48574.96 \ secs$ / $809.58 \ mins$ / $13.49 \ hours$	\\
					\hline
				\end{tabular}
			\end{sc}
	\end{center}
	\vskip -0.1in
\end{table*} 

Several observations can be made here: firstly, with the help of \textit{LIBSKYLARK}, our framework \ac{*SHL} can solve large data sets like \textit{Mnist1M}. As reported in \tref{table1}, \ac{*SHL} provides competitive retrieval results using about $13$ hours for the $45$ bits codeword. Secondly, as shown in the second and third row in the table, a larger cluster will benefit more from parallel computing. Here, \textit{Mnist} run on the $10$ nodes cluster need almost $25$\% running time compared with the same data set on a $2$ node cluster. Thus, when confronting with even larger data sets, a larger cluster with more powerful nodes can be a solution.

\acresetall

\section{Conclusions}
\label{sec:Conclusions}

In this paper we considered a novel hash learning framework, namely \ac{*SHL}. The method has the following main advantages: first, its \ac{MM}/\ac{BCD} training algorithm is efficient and simple to implement. Secondly, the framework is able to address supervised, unsupervised or even semi-supervised learning tasks. Additionally, after introducing a regularization over the multiple codewords, we also provide the \ac{PSD} method to solve this regularization.

In order to show the merits of the methods, we performed a series of experiments involving $5$ benchmark datasets. In experiments that were conducted, a comparison between our methods with the other $6$ state-of-the-art hashing methods shows \ac{*SHL} to be highly competitive. Moreover, we also give results on transductive learning scenario. Additionally, another application based on our framework, interactive image segmentation, is also showcased in the experimental section. Finally, we introduce \ac{*SHL} can also solve problems containing a huge number of samples.

\appendices

\section{Proof of \propref{prop:2.5}}
\label{proofprop3}


\begin{proof}
By replacing hinge function in \pref{eq:6}, we got the following problem for the first block minimization:

\begin{align}
\label{eq:s1}
\underset{\xi^b_{c,n,s}}{\underset{w_{b,m}, \beta_b }{min}} & \ \ \lambda_1 \sum_c \sum_s \sum_n \gamma'_{c,n, s} \xi^b_{c,n, s} + \frac{1}{2} \sum_m \frac{\left\| w_{b,m} \right\|^2_{\mathcal{H}_m}}{\theta_{b,m}} \nonumber \\
\text{s.t.} & \ \ \xi^b_{c,n, s} \geq 0 \nonumber \\
	 & \ \ \xi^b_{c,n, s} \geq 1-(\sum_m \left \langle w_{b,m} , \phi_m(x)  \right \rangle_{\mathcal{H}_m} + \beta_b)\mu^b_{c,s}
\end{align}

\noindent First of all, after considering Representer Theorem \cite{Scholkopf2001}, we have:

\begin{align}
\label{eq:s2}
w_{b,m} = \theta_{b,m}\sum_n \eta_{b,n} \phi_m(x_n)
\end{align}

\noindent where $n$ is index of the training samples. By defining $\boldsymbol{\xi}_b \in \mathbb{R}^{NCS}$ to be the vecotr containing all $\xi_{c,n,s}^b$'s, $\boldsymbol{\eta}_b \triangleq [\eta_{b,1},\eta_{b,2},...,\eta_{b,N}]^T \in \mathbb{R}^N$ and $\boldsymbol{\mu}_b \triangleq [\mu^b_{1,1}, \ldots, \mu^b_{1,S}, \mu^b_{2,1}, \ldots, \mu^b_{C,S}]^T \in \mathbb{R}^{CS}$, the vectorized version of \pref{eq:s1} with \eref{eq:s2}:

\begin{align}
\label{eq:s3}
\underset{\boldsymbol{\eta}_b, \boldsymbol{\xi}_b, \beta_b}{min} & \ \ \lambda_1 \boldsymbol{\gamma}' \boldsymbol{\xi}_b + \frac{1}{2} \boldsymbol{\eta}^T_b \mathbf{K}_b \boldsymbol{\eta}_b\nonumber \\
s.t. & \ \ \boldsymbol{\xi}_b \succeq \boldsymbol{0} \nonumber \\
	 & \ \ \boldsymbol{\xi}_b \succeq \boldsymbol{1}_{NCS}-(\boldsymbol{\mu}_b \otimes \mathbf{K}_b)\boldsymbol{\eta}_b-(\boldsymbol{\mu}_b \otimes \boldsymbol{1}_N)\beta_b
\end{align}

\noindent Where $\boldsymbol{\gamma}'$ and $\mathbf{K}_b$ are defined in \propref{prop:2.5}. Take the Lagrangian $\mathcal{L}$ and its derivatives, we have the following relations, here $\boldsymbol{\alpha}_b$ and $\boldsymbol{\zeta}_b$ are Lagrangian multipliers for the two constraints:

\begin{align}
\label{eq:s4}
\frac{\partial \mathcal{L}}{\partial \boldsymbol{\xi}_b} = \boldsymbol{0} & \Rightarrow \begin{cases}
\boldsymbol{\zeta}_b = \lambda_1 \boldsymbol{\gamma}'-\boldsymbol{\alpha}_b  \\ 
 \boldsymbol{0} \preceq \boldsymbol{\alpha}_b \preceq \lambda_1 \boldsymbol{\gamma}'
\end{cases} \\
\label{eq:s5}
\frac{\partial \mathcal{L}}{\partial \beta_b} = 0 & \Rightarrow \boldsymbol{\alpha}_b^T(\boldsymbol{\mu}_b \otimes\boldsymbol{1}_N) = 0 \\
\label{eq:s6}
\frac{\partial \mathcal{L}}{\partial \boldsymbol{\eta}_b} = \boldsymbol{0} & \overset{\exists \mathbf{K}_b^{-1}}{\Rightarrow} \boldsymbol{\eta}_b = \mathbf{K}_b^{-1}(\boldsymbol{\mu}_b \otimes \mathbf{K}_b)^T \boldsymbol{\alpha}_b  
\end{align}

\noindent Substitute \eref{eq:s4}, \eref{eq:s5} and \eref{eq:s6} back into $\mathcal{L}$, meanwhile, we notice the quatric term becomes: 

\begin{align}
\label{eq:s7}
& (\boldsymbol{\mu}_b \otimes \mathbf{K}_b) \mathbf{K}_b^{-1} (\boldsymbol{\mu}_b^T \otimes \mathbf{K}_b) \nonumber \\
= & (\boldsymbol{\mu}_b \otimes \mathbf{K}_b) (1 \otimes \mathbf{K}_b^{-1}) (\boldsymbol{\mu}_b^T \otimes \mathbf{K}_b) \nonumber \\
= & (\boldsymbol{\mu}_b \otimes \mathbf{I}_{N \times N})(\boldsymbol{\mu}_b^T \otimes \mathbf{K}_b) \nonumber \\
= & (\boldsymbol{\mu}_b \boldsymbol{\mu}_b^T) \otimes \mathbf{K}_b
\end{align}

\noindent \eref{eq:s7} can be further derived:

\begin{align}
\label{eq:s8}
& (\boldsymbol{\mu}_b \boldsymbol{\mu}_b^T) \otimes \mathbf{K}_b \nonumber \\
& = [(\diag{\boldsymbol{\mu}_b} \boldsymbol{1}_C)(\diag{\boldsymbol{\mu}_b} \boldsymbol{1}_C)^T] \otimes \mathbf{K}_b \nonumber \\
& = [\diag{\boldsymbol{\mu}_b}(\boldsymbol{1}_C \boldsymbol{1}_C^T)\diag{\boldsymbol{\mu}_b}] \otimes [\mathbf{I}_N \mathbf{K}_b \mathbf{I}_N] \nonumber \\
& = [\diag{\boldsymbol{\mu}_b} \otimes \mathbf{I}_N][(\boldsymbol{1}_C \boldsymbol{1}_C^T) \otimes \mathbf{K}_b][\diag{\boldsymbol{\mu}_b} \otimes \mathbf{I}_N] \nonumber \\
& = [\diag{\boldsymbol{\mu}_b \otimes \boldsymbol{1}_N}][(\boldsymbol{1}_C \boldsymbol{1}_C^T) \otimes \mathbf{K}_b][\diag{\boldsymbol{\mu}_b \otimes \boldsymbol{1}_N}] \nonumber \\
& = \mathbf{D}_b [(\boldsymbol{1}_C \boldsymbol{1}_C^T) \otimes \mathbf{K}_b] \mathbf{D}_b
\end{align}

\noindent The first equality comes from $\diag{\boldsymbol{u}}\boldsymbol{1} = \boldsymbol{u}$ for some vector $\boldsymbol{u}$. The third equality is the mixed-product property of Kronecker product. This relation $\diag{\boldsymbol{u} \otimes \boldsymbol{1}} = \diag{\boldsymbol{u}} \otimes \mathbf{I}$ gives the fourth equality. $\mathbf{D}_b$ is defined in \propref{prop:2.5}.

After considering \eref{eq:s7} and \eref{eq:s8}, we get the final dual form as shown in \propref{prop:2.5}.
\end{proof}

\section{Proof of \propref{prop:3.5}}
\label{proofprop4}

\begin{proof}
With the definitions of proximal operator \eref{eq:3_6}, we have the following problem to minimize over:

\begin{align}
\label{eq:s9}
P(\boldsymbol{\mu}) & = \eta \left \| U \boldsymbol{\mu} \right \|_2 + \frac{1}{2} \left \| \boldsymbol{v} - \boldsymbol{\mu} \right \|^2_2 \nonumber \\
& = \eta \left \| U \boldsymbol{\mu} \right \|_2 + \frac{1}{2} \left \| \boldsymbol{v}_1 - \boldsymbol{\mu}_1 \right \|^2_2 + \cdot \cdot \cdot + \frac{1}{2} \left \| \boldsymbol{v}_S - \boldsymbol{\mu}_S \right \|^2_2
\end{align}

\noindent
The second equality follows from the definitions of the vectors $\boldsymbol{\mu}$ and $\boldsymbol{v}$. Since $L_2$ norm is non differentiable at point $0$, we optimize \eref{eq:s9} in two cases.

\textbf{Case 1:} when $\boldsymbol{\mu}_i \neq \boldsymbol{\mu}_j$, we take the gradients for each individual $\boldsymbol{\mu}_1$ to $\boldsymbol{\mu}_S$:

\begin{align}
\label{eq:s10}
\begin{cases}
 \frac{\partial P(\boldsymbol{\mu})}{\partial \boldsymbol{\mu}_1}  = \boldsymbol{\mu}_1 - \boldsymbol{v}_1 = \boldsymbol{0} \\ 
 \quad \quad \ \ \vdots \\ 
 \frac{\partial P(\boldsymbol{\mu})}{\partial \boldsymbol{\mu}_i} = \eta \frac{\boldsymbol{\mu}_i - \boldsymbol{\mu}_j}{\left \| \boldsymbol{\mu}_i - \boldsymbol{\mu}_j \right \|_2} + \boldsymbol{\mu}_i - \boldsymbol{v}_i = \boldsymbol{0} \\ 
 \quad \quad \ \ \vdots \\ 
 \frac{\partial P(\boldsymbol{\mu})}{\partial \boldsymbol{\mu}_j} = \eta \frac{\boldsymbol{\mu}_j - \boldsymbol{\mu}_i}{\left \| \boldsymbol{\mu}_i - \boldsymbol{\mu}_j \right \|_2} + \boldsymbol{\mu}_j - \boldsymbol{v}_j = \boldsymbol{0} \\ 
 \quad \quad \ \ \vdots \\
 \frac{\partial P(\boldsymbol{\mu})}{\partial \boldsymbol{\mu}_S}  = \boldsymbol{\mu}_S - \boldsymbol{v}_S = \boldsymbol{0}
\end{cases}
\end{align}

\noindent
Solve the linear equations with $\boldsymbol{\mu}_i$ and $\boldsymbol{\mu}_j$:

\begin{align}
\label{eq:s11}
\begin{cases}
 \boldsymbol{\mu}_i = \frac{1}{2\tau + 1} \left[ (1 + \tau)\boldsymbol{v}_i + \tau \boldsymbol{v}_j \right ]\\
 \boldsymbol{\mu}_j = \frac{1}{2\tau + 1} \left[ \tau \boldsymbol{v}_i + (1 + \tau) \boldsymbol{v}_j \right ]
\end{cases}
\end{align}

\noindent
where $\tau \triangleq \eta / \delta$ and $\delta \triangleq \left \| \boldsymbol{\mu}_i - \boldsymbol{\mu}_j \right \|_2$. Now we have the following derivations:

\begin{align}
\label{eq:s12}
& \boldsymbol{\mu}_i - \boldsymbol{\mu}_j = \frac{1}{2\tau + 1} (\boldsymbol{v}_i - \boldsymbol{v}_j) \nonumber \\
& \Rightarrow \left \| \boldsymbol{\mu}_i - \boldsymbol{\mu}_j \right \|_2 = \frac{1}{2\tau + 1} \left \| \boldsymbol{v}_i - \boldsymbol{v}_j \right \|_2 \nonumber \\
& \Rightarrow \delta = \frac{\left \| \boldsymbol{v}_i - \boldsymbol{v}_j \right \|_2}{2\frac{\eta}{\delta} + 1} \nonumber \\ 
& \Rightarrow \delta = \left \| \boldsymbol{\mu}_i - \boldsymbol{\mu}_j \right \|_2 =  \left \| \boldsymbol{v}_i - \boldsymbol{v}_j \right \|_2 - 2\eta  
\end{align}

\noindent
Plug $\tau$ and \eref{eq:s12} into \eref{eq:s11}, we achieve the results for $\boldsymbol{\mu}_i$ and $\boldsymbol{\mu}_j$:

\begin{align}
\label{eq:s13}
\begin{cases}
\boldsymbol{\mu}_i = \alpha_1 \boldsymbol{v}_i + \alpha_2 \boldsymbol{v}_j \\ 
\boldsymbol{\mu}_j = \alpha_2 \boldsymbol{v}_i + \alpha_1 \boldsymbol{v}_j
\end{cases}
\end{align}

\noindent
Here $\alpha_1 = 1 - \alpha_2$ and $\alpha_2 = \frac{\eta}{\left \| \boldsymbol{v}_i - \boldsymbol{v}_j \right \|_2}$. Additionally, \eref{eq:s12} is larger than $0$ which gives the following condition for \textbf{Case 1}: $0 < \eta \leq \frac{\left \| \boldsymbol{v}_i - \boldsymbol{v}_j \right \|_2}{2}$.

\textbf{Case 2:} when $\boldsymbol{\mu}_i = \boldsymbol{\mu}_j$, $\boldsymbol{\mu}_i$ and $\boldsymbol{\mu}_j$ are represented as:

\begin{align}
\label{eq:s14}
\begin{cases}
\boldsymbol{\mu}_i = \frac{1}{2} \boldsymbol{v}_i + \frac{1}{2} \boldsymbol{v}_j \\ 
\boldsymbol{\mu}_j = \frac{1}{2} \boldsymbol{v}_i + \frac{1}{2} \boldsymbol{v}_j
\end{cases}
\end{align}

\noindent
Note that this case satisfies when $\eta > \frac{\left \| \boldsymbol{v}_i - \boldsymbol{v}_j \right \|_2}{2}$. 

Combining two cases, the results provided in \propref{prop:3.5} are achieved.

\end{proof}


\section{Proof of \lemmaref{lemma:1}}
\label{prooflemma1}

\begin{proof}
From \defref{def:def2}, we have:

\begin{align}
\label{eq:lem1-1}
& \widehat{\Re}_Q \left( \Psi \circ \widetilde{\mathcal{F}} \right) = \frac{1}{N} \Es{\sigma}{\sup_{\boldsymbol{f} \in \boldsymbol{\widetilde{F}}} \sum_n \sigma_n \Psi(\boldsymbol{f}(z_n))} = \nonumber \\
& = \frac{1}{N}\Es{\boldsymbol{\sigma}_{N-1}}{\Es{\sigma_N}{\sup_{\boldsymbol{f} \in \widetilde{\mathcal{F}}} \left[u(\boldsymbol{f}) + \sigma_N\Psi(\boldsymbol{f}(z_N)) \right]}} \nonumber \\
& = \frac{1}{N}\Es{\boldsymbol{\sigma}_{N-1}}{A(\boldsymbol{\sigma}_{N-1})}
\end{align}

\noindent
where $u(\boldsymbol{f}) \triangleq \sum_{n = 1}^{N-1}\sigma_n \Psi(\boldsymbol{f}(z_n))$ and $A(\boldsymbol{\sigma}_{N-1}) \triangleq \Es{\sigma_N}{\sup_{\boldsymbol{f} \in \widetilde{\mathcal{F}}} \left[u(\boldsymbol{f}) + \sigma_N\Psi(\boldsymbol{f}(z_N)) \right]}$.

Expanding the expectation , we get:

\begin{align}
\label{eq:lem1-2}
A(\boldsymbol{\sigma}_{N-1}) = \frac{1}{2} [ & \sup_{\boldsymbol{f} \in \widetilde{\mathcal{F}}} \left[ u(\boldsymbol{f}) + \Psi(\boldsymbol{f}(z_n)) \right] \nonumber \\
& + \sup_{\boldsymbol{f} \in \widetilde{\mathcal{F}}} \left[ u(\boldsymbol{f}) - \Psi(\boldsymbol{f}(z_n)) \right] ]
\end{align}

\noindent
Additionally, we define the following: $\widehat{B}(\boldsymbol{f}) \triangleq u(\boldsymbol{f}) + \Psi(\boldsymbol{f}(z_n))$ and $\widetilde{B}(\boldsymbol{f}) \triangleq u(\boldsymbol{f}) - \Psi(\boldsymbol{f}(z_n))$.

From the superium's definition, we have that $\forall \epsilon > 0$, there are $\boldsymbol{\widehat{f}}$ amd $\boldsymbol{\widetilde{f}}$ in $\mathcal{\widetilde{F}}$ such that:

\begin{align}
\label{eq:lem1-3}
\sup_{\boldsymbol{f} \in \widetilde{\mathcal{F}}} \widehat{B}(\boldsymbol{f}) \geq \widehat{B}(\boldsymbol{\widehat{f}}) \geq (1 - \epsilon) \sup_{\boldsymbol{f} \in \widetilde{\mathcal{F}}} \widehat{B}(\boldsymbol{f}) \\
\label{eq:lem1-4}
\sup_{\boldsymbol{f} \in \widetilde{\mathcal{F}}} \widetilde{B}(\boldsymbol{f}) \geq \widetilde{B}(\boldsymbol{\widehat{f}}) \geq (1 - \epsilon) \sup_{\boldsymbol{f} \in \widetilde{\mathcal{F}}} \widetilde{B}(\boldsymbol{f})
\end{align}

From \eref{eq:lem1-3} and \eref{eq:lem1-4}, for any $\epsilon > 0$, we have:

\begin{align}
\label{eq:lem1-5}
& (1 - \epsilon)A(\boldsymbol{\sigma}_{N-1}) \leq \frac{1}{2} \left[ \widehat{B}(\boldsymbol{\widehat{f}}) + \widetilde{B}(\boldsymbol{\widetilde{f}}) \right] = \nonumber \\ 
& = \frac{1}{2} \left[ u(\boldsymbol{\widehat{f}}) + u(\boldsymbol{\widetilde{f}}) + \Psi(\boldsymbol{\widehat{f}}(z_n)) - \Psi(\boldsymbol{\widetilde{f}}(z_n)) \right]
\end{align}

Since $\Psi$ is $L$-Lipschitz continuous w.r.t the $\left \| \cdot \right \|_1$ norm, it holds that:

\begin{align}
\label{eq:lem1-6}
\Psi(\boldsymbol{\widehat{f}}(z_n)) & - \Psi(\boldsymbol{\widetilde{f}}(z_n)) \leq L\left \| \boldsymbol{\widehat{f}}(z_n) - \boldsymbol{\widetilde{f}}(z_n) \right \|_1 = \nonumber \\
& = L\sum_{b = 1}^{B} | \widehat{f}_b(z_n) - \widetilde{f}_b(z_n) | = \nonumber \\
& = L \sum_{b = 1}^{B} q_b\left( \widehat{f}_b(z_n) - \widetilde{f}_b(z_n) \right)
\end{align}

\noindent
where $q_b \triangleq sgn \left( \widehat{f}_b(z_n) - \widetilde{f}_b(z_n) \right)$. From \eref{eq:lem1-5} and \eref{eq:lem1-6}, we obtain:

\begin{align}
\label{eq:lem1-7}
& (1 - \epsilon) A(\boldsymbol{\sigma}_{N - 1}) \nonumber \\
& \leq \frac{1}{2} \left[ u(\boldsymbol{\widehat{f}}) + u(\boldsymbol{\widetilde{f}}) + L \sum_{b = 1}^B q_b\left( \widehat{f}_b(z_n) - \widetilde{f}_b(z_n) \right) \right] =  \nonumber \\
& = \frac{1}{2} \left[ \left[ u(\boldsymbol{\widehat{f}}) + L\sum_{b = 1}^B q_b \widehat{f}_b(z_N) \right] + \left[ u(\boldsymbol{\widetilde{f}}) - L\sum_{b = 1}^B q_b \widetilde{f}_b(z_N) \right] \right] \nonumber \\
\end{align}

\noindent
By the definition of superium, \eref{eq:lem1-7} is bounded by:

\begin{align}
\label{eq:lem1-8}
 \eeref{eq:lem1-7} \leq \sup_{\boldsymbol{q} \in \mathbb{H}^B} \frac{1}{2} \bigg[ \bigg[ u(\boldsymbol{\widehat{f}}) + L \sum_b^B & q_b \widehat{f}_b(z_N) \bigg] + \bigg[ u(\boldsymbol{\widetilde{f}}) \nonumber \\
 & - L \sum_b^B q_b \widetilde{f}(z_N) \bigg]\bigg] \leq \nonumber \\
 \leq \frac{1}{2} \bigg[ \sup_{\boldsymbol{q} \in \mathbb{H}^B} \bigg[ u(\boldsymbol{\widehat{f}}) + L \sum_b^B & q_b \widehat{f}(z_N) \bigg] \bigg] + \sup_{\boldsymbol{q} \in \mathbb{H}^B} \bigg[ u(\boldsymbol{\widetilde{f}}) \nonumber \\
 & - L\sum_b^B q_b \widetilde{f}_b(z_N)\bigg] \bigg] 
\end{align}

\noindent
With the help of \eref{eq:lem1-3} and \eref{eq:lem1-4}, \eref{eq:lem1-8} is bounded:

\begin{align}
\label{eq:lem1-9}
\eeref{eq:lem1-8} & \leq \frac{1}{2} \bigg[ \sup_{\boldsymbol{q} \in \mathbb{H}^B} \sup_{\boldsymbol{f} \in \mathcal{\widetilde{F}}} \bigg[ u(\boldsymbol{f}) + L \sum_b^B q_bf_b(z_N) \bigg] + \nonumber \\
& \sup_{\boldsymbol{q} \in \mathbb{H}^B} \sup_{\boldsymbol{f} \in \mathcal{\widetilde{F}}} \bigg[ u(\boldsymbol{f}) - L\sum_b^B q_b f_b(z_N) \bigg] \bigg] = \nonumber \\
& = \Es{\sigma_N}{\sup_{\boldsymbol{q} \in \mathbb{H}^B} \sup_{\boldsymbol{f} \in \mathcal{\widetilde{F}}} \left[ u(\boldsymbol{f}) + \sigma_N L \sum_b^B q_b f_b (z_N) \right]} = \nonumber \\
& = \Es{\sigma_N}{\sup_{\boldsymbol{f} \in \mathcal{\widetilde{F}}} \left[ u(\boldsymbol{f}) + \sigma_NL \sup_{\boldsymbol{q} \in \mathbb{H}^B} \sum_b^B q_bf_b(z_N) \right]} = \nonumber \\
& = \Es{\sigma_N}{\sup_{\boldsymbol{f} \in \mathcal{\widetilde{F}}} \left[ u(\boldsymbol{f}) + \sigma_NL \sum_b^B \sup_{q_b \in \{\pm 1 \}}  q_bf_b(z_N) \right]} =  \nonumber \\
& = \Es{\sigma_N}{\sup_{\boldsymbol{f} \in \mathcal{\widetilde{F}}} \left[ u(\boldsymbol{f}) + \sigma_NL \sum_b^B sgn(f_b(z_N)) f_b(z_N) \right]} = \nonumber \\
& = \Es{\sigma_N}{\sup_{\boldsymbol{f} \in \mathcal{\widetilde{F}}} \left[ u(\boldsymbol{f}) + \sigma_NL \left \| \boldsymbol{f}(z_N) \right \|_1 \right]}
\end{align}

\noindent
Since \eref{eq:lem1-9} holds for every $\epsilon > 0$, we have that :

\begin{align}
\label{eq:lem1-10}
A(\boldsymbol{\sigma}_{N-1}) \leq \Es{\sigma_N}{\sup_{\boldsymbol{f} \in \mathcal{\widetilde{F}}} \left[ u(\boldsymbol{f}) + \sigma_NL \left \| \boldsymbol{f}(z_N) \right \|_1 \right]}
\end{align}

\noindent
Repeating this process for the remaining $\sigma$ eventually yields the result of this lemma.
\end{proof}

\section{Proof of \lemmaref{lemma:2}}
\label{prooflemma2}

\begin{proof}
Let $\Psi(\cdot) \triangleq \left \| \cdot \right \|_1$. Utilizing the similar technique in \lemmaref{lemma:1}, by defining $u(\boldsymbol{f}) \triangleq \sum_{n = 1}^{N - 1} \sigma_N \Psi(\boldsymbol{f}(z_N))$, we have:

\begin{align}
\label{eq:lem2-1}
\widehat{\Re}_Q(\left \| \mathcal{\widetilde{F}} \right \|_1) = \frac{1}{N} \Es{\boldsymbol{\sigma}_{N-1}}{A(\boldsymbol{\sigma}_{N-1})}
\end{align}

\noindent
Here $A(\boldsymbol{\sigma}_{N-1}) = \Es{\sigma_N}{\sup_{\boldsymbol{f} \in \mathcal{\widetilde{F}}} \left[ u(\boldsymbol{f}) + \sigma_N \Psi(\boldsymbol{f}(z_N)) \right]}$. Similarly, by defining $\widehat{B}(\boldsymbol{f}) \triangleq u(\boldsymbol{f}) + \Psi(\boldsymbol{f}(z_N))$ and $\widetilde{B}(\boldsymbol{f}) \triangleq u(\boldsymbol{f}) - \Psi(\boldsymbol{f}(z_N))$, we have for any $\epsilon > 0$:

\begin{align}
\label{eq:lem2-5}
& (1 - \epsilon)A(\boldsymbol{\sigma}_{N - 1}) \leq \frac{1}{2} \left[ \widehat{B}(\boldsymbol{\widehat{f}}) + \widetilde{B}(\boldsymbol{\widetilde{f}})\right] = \nonumber \\
& = \frac{1}{2} \left[ u(\boldsymbol{\widehat{f}}) + u(\boldsymbol{\widetilde{f}}) + \Psi(\boldsymbol{\widehat{f}}(z_N)) - \Psi(\boldsymbol{\widetilde{f}}(z_N)) \right]
\end{align}

\noindent
By the reverse triangle inequality and $|\cdot|$'s $1$ - Lipschitz property:

\begin{align}
\label{eq:lem2-6}
& \Psi(\boldsymbol{\widehat{f}}(z_N)) - \Psi(\boldsymbol{\widetilde{f}}(z_N)) = \sum_{b=1}^B\left[ |\widehat{f}_b(z_N)| - |\widetilde{f}_b(z_N)| \right] \leq \nonumber \\
& \leq \sum_{b = 1}^B sgn(\widehat{f}_b(z_N) - \widetilde{f}_b(z_N))(\widehat{f}_b(z_N) - \widetilde{f}_b(z_N))
\end{align}

\noindent
With the definition of $q_b \triangleq sgn(\widehat{f}_b(z_N) - \widetilde{f}_b(z_N))$, we combine \eref{eq:lem2-5} and \eref{eq:lem2-6}:

\begin{align}
\label{eq:lem2-7}
& (1 - \epsilon) A(\boldsymbol{\sigma}_{N - 1}) \nonumber \\
& \leq \frac{1}{2} \left[ u(\boldsymbol{\widehat{f}}) + u(\boldsymbol{\widetilde{f}}) + \sum_{b = 1}^B q_b(\widehat{f}_b(z_N) - \widetilde{f}_b(z_N)) \right] = \nonumber \\
& = \frac{1}{2} \left[ \left[ u(\boldsymbol{\widehat{f}}) + \sum_{b = 1}^B q_b \widehat{f}_b(z_N) \right] + \left[ u(\boldsymbol{\widetilde{f}}) - \sum_{b = 1}^B q_b \widetilde{f}_b(z_N) \right] \right]
\end{align}

\noindent
For $b \in \mathbb{H}_B$, define $q'_b \triangleq q_b$ if $q_b \neq 0$ and $q'_b \triangleq 1$ otherwise. Also, define $\widehat{f}'(\cdot) \triangleq q'_b \widehat{f}(\cdot)$, then we have $\widehat{f}(\cdot) = q'_b\widehat{f}'(\cdot)$ and :

\begin{align}
\label{eq:lem2-8}
& u(\boldsymbol{\widehat{f}}) + \sum_{b = 1}^B q_b \widehat{f}_b(z_N) = \nonumber \\
& = u\left( q'_1\widehat{f}'_1(\cdot), ..., q'_B \widehat{f}'_B(\cdot) \right) + \sum_{b = 1}^B \widehat{f}'_b(z_N) = \nonumber \\
& = \sum_{n = 1}^{N - 1} \sigma_n \sum_{b = 1}^B |q'_b \widehat{f}'_b(z_n)| + \sum_{b = 1}^B \widehat{f}'_b(z_N) = \nonumber \\
& = u(\boldsymbol{\widehat{f}}') + \sum_{b = 1}^B \widehat{f}'_b(z_N) \leq \sup_{\boldsymbol{f} \in \mathcal{\widetilde{F}}} \left[ u(\boldsymbol{f}) + \sum_{b = 1}^B f_b(z_N) \right]
\end{align}

\noindent
The above derivation is based on the fact that if $\left[ f_1(z), ..., f_B(z) \right]^T \in \mathcal{\widetilde{F}}$, then we also have $\left[ \pm f_1(z),...,f_B(z) \right]^T \in \mathcal{\widetilde{F}}$ for $z \in \mathcal{Z}$. 

Using a similar rationale, we can show that:

\begin{align}
\label{eq:lem2-9}
u(\boldsymbol{\widetilde{f}}) - \sum_{b = 1}^B q_b \widetilde{f}_b(z_n) \leq \sup_{\boldsymbol{f} \in \mathcal{\widetilde{F}}} \left[ u(\boldsymbol{f}) - \sum_{b = 1}^B f_b(z_N) \right]
\end{align}

\noindent
Combine \eref{eq:lem2-7}, \eref{eq:lem2-8} and \eref{eq:lem2-9}:

\begin{align}
\label{eq:lem2-10}
& (1 - \epsilon) A(\boldsymbol{\sigma}_{N - 1}) \nonumber \\
& \leq \frac{1}{2} \left[ \sup_{\boldsymbol{f} \in \mathcal{\widetilde{F}}} \left[ u(\boldsymbol{f}) + \sum_{b = 1}^B f_b(z_N) \right] + \sup_{\boldsymbol{f} \in \mathcal{\widetilde{F}}} \left[ u(\boldsymbol{f}) - \sum_{b = 1}^B f_b(z_N) \right]\right] \nonumber \\
& = \Es{\sigma_N}{\sup_{\boldsymbol{f} \in \mathcal{\widetilde{F}}} \left[ u(\boldsymbol{f}) + \sigma_N \sum_{b = 1}^B f_b(z_N ) \right]}
\end{align}

\noindent
Since \eref{eq:lem2-10} holds for every $\epsilon > 0$, we have that:

\begin{align}
\label{eq:lem2-11}
A(\boldsymbol{\sigma}_{N-1}) \leq \Es{\sigma_N}{\sup_{\boldsymbol{f} \in \mathcal{\widetilde{F}}} \left[ u(\boldsymbol{f}) + \sigma_N \boldsymbol{1}^T \boldsymbol{f}(z_N) \right] }
\end{align}

\noindent
Repeating this process for the remaining $\sigma$s will eventually yield the result of this lemma.

\end{proof}


\section{Proof of \thmref{theorem1}}
\label{prooftheorem}
\begin{proof}
Consider the function spaces:

\begin{align}
\mathcal{G} \triangleq  \{ \boldsymbol{g}:(x,l) \mapsto \left[ \mu_1(l)f_1(x), ..., \mu_B(l) f_B(x) \right]^T, \  \nonumber \\
 \boldsymbol{\mu} \in \mathcal{M}, \ \boldsymbol{f}: \mathcal{X} \mapsto \mathbb{R}^B \} \nonumber
\end{align} 
 
\begin{align}
\Psi \circ \mathcal{G} \triangleq \left \{ \Psi(\boldsymbol{g}(\cdot)):(x,l) \mapsto \frac{1}{B} \sum_{b = 1}^B \Phi_{\rho}(g_b(x,l)), \ \boldsymbol{g} \in \mathcal{G} \right \} \nonumber
\end{align}

\noindent
Notice that, since $\Phi_{\rho}(u) \in [0,1]$ for $\forall u \in \mathbb{R}$, also $\Psi(\boldsymbol{g}(x,l)) \in [0,1]$ for $\forall \boldsymbol{g} \in \mathcal{G}$, $x \in \mathcal{x}$ and $l \in \mathbb{N}_G$. Hence, from Theorem 3.1 of \cite{Mohri2012}, for fixed (independent of $Q$) $\rho > 0$ and for any $\delta > 0$ and any $\boldsymbol{g} \in \mathcal{G}$, with probability at least $1 - \delta$, it holds that:

\begin{align}
\label{eq:thm3-1}
\mathbb{E} \left \{ \Psi(\boldsymbol{g}(x,l)) \right \} \leq \widehat{\mathbb{E}}_Q \left \{ \Psi(\boldsymbol{g}(x,l)) \right \} + 2\Re_N(\Psi \circ \mathcal{G}) + \sqrt{\frac{ln\frac{1}{\delta}}{2N}}
\end{align}

\noindent
Where we define $\forall h: \mathcal{X} \times \mathbb{N}_C \mapsto \mathbb{R}$ and $\widehat{\mathbb{E}}_Q \left\{ h(x, l) \right \} \triangleq \frac{1}{N} \sum_{n=1}^N h(x_n, l_n)$. Since $[u < 0] \leq \Phi_{\rho}(u)$ for $\forall u \in \mathbb{R}, \ \rho > 0$, it holds that:

\begin{align}
& \frac{1}{B}d \left( \sgn \boldsymbol{f}(x), \boldsymbol{\mu}(l) \right ) = \frac{1}{B}\sum_{b = 1}^B \left[ \mu_b(l)f_b(x) < 0 \right] \leq \nonumber \\
& \leq \frac{1}{B} \sum_{b=1}^B \Phi_{\rho} \left( \mu_b(l)f_b(x) \right ) = \Psi(\boldsymbol{g}(x,l)) \nonumber
\end{align}

\noindent
Thus, we have the following:

\begin{align}
\label{eq:thm3-2}
er(\boldsymbol{f}, \boldsymbol{\mu}) \triangleq \mathbb{E} \left \{ \frac{1}{B}d(\sgn \boldsymbol{f}(x), \boldsymbol{\mu}(l)) \right \} \leq \mathbb{E} \left \{ \Psi(\boldsymbol{g}(x,l)) \right \}
\end{align}

\noindent
Due to \eref{eq:thm3-2}, with probability at least $1 - \delta$, \eref{eq:thm3-1} becomes now:

\begin{align}
\label{eq:thm3-3}
er(\boldsymbol{f}, \boldsymbol{\mu}) \leq \widehat{\mathbb{E}}_Q \left \{ \Psi(\boldsymbol{g}(x,l)) \right \} + 2 \Re_N(\Psi \circ \mathcal{G}) + \sqrt{\frac{ln\frac{1}{\delta}}{2N}}
\end{align}

\noindent
Now due to the fact that $\Psi(\cdot)$ is $\frac{1}{B\rho}$ - Lipschitz continuous w.r.t $\left \| \cdot \right \|_1$ and \lemmaref{lemma:1}, we have:

\begin{align}
\widehat{\Re}_Q \left(\Psi \circ \mathcal{G}\right) \leq \frac{1}{B\rho}\widehat{\Re}_Q \left( \left \| \mathcal{G} \right \|_1 \right) \nonumber
\end{align}

\noindent
Also, since $\boldsymbol{\mu}: \mathbb{N}_G \mapsto \mathbb{H}^B$, we have $\widehat{\Re}_Q \left( \left \| \mathcal{G} \right \|_1 \right) = \widehat{\Re}_Q \left( \left \| \bar{\mathcal{F}} \right \|_1 \right)$, where $\bar{\mathcal{F}}$ is defined in \eref{eq:4-2}, we get:

\begin{align}
\widehat{\Re}_Q\left( \Psi \circ \mathcal{G} \right) \leq \frac{1}{B\rho} \widehat{\Re}_Q\left( \left \| \bar{\mathcal{F}} \right \|_1 \right) \nonumber
\end{align}

\noindent
Now due to \lemmaref{lemma:2}, we have $\widehat{\Re}_Q\left( \left \| \bar{\mathcal{F}} \right \|_1 \right) \leq \widehat{\Re}_Q \left( \boldsymbol{1}^T \bar{\mathcal{F}} \right)$, by taking expectations on both sides w.r.t Q, the above inequality becomes:

\begin{align}
\label{eq:thm3-4}
\Re_N\left( \Psi \circ \mathcal{G} \right) \leq \frac{1}{B\rho} \Re_N \left( \boldsymbol{1}^T \bar{\mathcal{F}} \right)
\end{align}

\noindent
Substitute \eref{eq:thm3-4} into \eref{eq:thm3-3}:

\begin{align}
\label{eq:thm3-5}
er(\boldsymbol{f}, \boldsymbol{\mu}) \leq \widehat{\mathbb{E}}_Q \{ \Psi(\boldsymbol{g}(x, l)) \} + \frac{2}{B\rho}\Re_N(\boldsymbol{1}^T \bar{\mathcal{F}}) + \sqrt{\frac{ln\frac{1}{\delta}}{2N}}
\end{align}

\noindent
From the optimization problem in \eref{eq:6}, we note that \ac{*SHL} is utilizing the hypothesis spaces defined in \eref{eq:4-2} and \eref{eq:4-3}. Note the fact that each hash function of \ac{*SHL} is determined by the data independent of the others.

By considering the Representer Theorem \cite{Scholkopf2001}, we have $w = \sum_{n = 1} \alpha_n \Phi_{\boldsymbol{\theta}}(x_n), \ \boldsymbol{\alpha} \in \mathbb{R}^N$, which implies: $f(x) = \sum_n \alpha_n k_{\boldsymbol{\theta}}(x, x_n) + \beta$ and $\left \| w \right \|^2_{\mathcal{H}_{\boldsymbol{\theta}}} = \boldsymbol{\alpha}^T \boldsymbol{K}_{\boldsymbol{\theta}} \boldsymbol{\alpha}$. Here $\boldsymbol{K}_{\boldsymbol{\theta}}$ is the kernel matrix of the training data.

Hence, $\mathcal{F}$ can be re-expressed as:

\begin{align}
\mathcal{F} = \bigg \{ f: x \mapsto \sum_n \alpha_n k_{\boldsymbol{\theta}}(x, x_n) + \beta, \ \beta \in \mathbb{R} \nonumber \\
, \ \boldsymbol{\alpha} \in \Omega_{\boldsymbol{\alpha}}(\boldsymbol{\theta}), \ \boldsymbol{\theta} \in \Omega_{\boldsymbol{\theta}} \bigg \} \nonumber 
\end{align}

\noindent
where $\Omega_{\boldsymbol{\alpha}}(\boldsymbol{\theta}) \triangleq \{ \boldsymbol{\alpha} \in \mathbb{R}^N: \ \boldsymbol{\alpha}^T \boldsymbol{K}_{\boldsymbol{\theta}} \boldsymbol{\alpha} \leq R^2 \}$.

First of all, let's upper bound the Rademacher Complexity of \ac{*SHL}'s hypothesis space:

\begin{align}
\label{eq:thm3-6}
& \widehat{\Re}_Q(\boldsymbol{1}^T \bar{\mathcal{F}}) = \frac{1}{N} \mathbb{E}_{\boldsymbol{\sigma}} \left \{ \sup_{\boldsymbol{f} \in \bar{\mathcal{F}}} \sum_n \sigma_n \sum_{b = 1}^B f_b(x_n) \right \} = \nonumber \\
& = \frac{1}{N} \mathbb{E}_{\boldsymbol{\sigma}} \left \{ \sup_{f_b \in \mathcal{F}, b \in \mathbb{N}_B } \sum_b \sum_n \sigma_n f_b(x_n) \right \} = \nonumber \\
& = \sum_b \frac{1}{N} \mathbb{E}_{\boldsymbol{\sigma}} \left \{ \sup_{f_b \in \mathcal{F}} \sum_n \sigma_n f_b(x_n) \right \} = B \widehat{\Re}_Q(\mathcal{F})
\end{align}

Next, we will upper bound $\widehat{\Re}_Q(\mathcal{F})$:

\begin{align}
\label{eq:thm3-7}
& \widehat{\Re}_Q(\mathcal{F}) = \frac{1}{N} \Es{\boldsymbol{\sigma}}{\sup_{f \in \mathcal{F}} \sum_n \sigma_n f(x_n)} = \nonumber \\
& = \Es{\boldsymbol{\sigma}}{\sup_{\boldsymbol{\alpha} \in \Omega_{\boldsymbol{\alpha}}(\boldsymbol{\theta}), \boldsymbol{\theta} \in \Omega_{\boldsymbol{\theta}}} \boldsymbol{\alpha}^T \boldsymbol{K}_{\boldsymbol{\theta}} \boldsymbol{\alpha} + \sup_{\beta \in \mathbb{R}} \sum_n \sigma_n \beta } \leq \nonumber \\
& \leq \frac{R}{N} \Es{\boldsymbol{\sigma}}{\sup_{\boldsymbol{\theta} \in \Omega_{\boldsymbol{\theta}}} \sqrt{\boldsymbol{\alpha}^T \boldsymbol{K}_{\boldsymbol{\theta}} \boldsymbol{\alpha}}} = \frac{R}{N} \Es{\boldsymbol{\sigma}}{ \sqrt{\sup_{\boldsymbol{\theta} \in \Omega_{\boldsymbol{\theta}}} \boldsymbol{\theta}^T \boldsymbol{u} }}
\end{align}

\noindent
where $\boldsymbol{u} \in \mathbb{R}^M$ such that $u_m \triangleq \boldsymbol{\sigma}^T K_m \boldsymbol{\alpha}$. The above inequality holds because of Cauchy-Schwarz inequality. Additionally, $\Es{\boldsymbol{\sigma}}{\sup_{\beta \in \mathbb{R}} \sum_n \sigma_n \beta} = 0$ since $\beta$ is bounded.

By the definition of the dual norm, if $p' \triangleq \frac{p}{p-1}$, we have:

\begin{align}
\label{eq:thm3-8}
\sup_{\boldsymbol{\theta} \in \Omega_{\boldsymbol{\theta}}} \boldsymbol{\theta}^T \boldsymbol{u} = \left \| \boldsymbol{u} \right \|_{p'}
\end{align}

Thus, \eref{eq:thm3-7} becomes:

\begin{align}
&\widehat{\Re}_Q(\mathcal{F}) \leq \frac{R}{N} \Es{\boldsymbol{\sigma}}{\sqrt{\left \| \boldsymbol{u} \right \|_{p'}}} = \nonumber \\ 
& = \frac{R}{N} \Es{\boldsymbol{\sigma}}{ \left[ \sum_{m = 1}^M (\boldsymbol{\sigma}^T K_m \boldsymbol{\sigma})^{p'}  \right]^{\frac{1}{2p'}}} \leq \nonumber \\ 
& \leq \frac{R}{N} \left[ \sum_m \Es{\boldsymbol{\sigma}}{(\boldsymbol{\sigma}^T K_m \boldsymbol{\sigma})^{p'}} \right]^{\frac{1}{2p'}} \nonumber
\end{align}

\noindent
The above inequality holds because of Jensen's Inequality. By the Lemma 5 from \cite{Li2015}, the above expression is upper bounded by:

\begin{align}
\label{eq:thm3-9}
& \frac{R}{N} \left[ \sum_m (p')^{\frac{p'}{2}} (\trace{K_m})^{\frac{p'}{2}} \right]^{\frac{1}{2p'}} = \nonumber \\
& = \frac{R}{N}(p')^{\frac{1}{4}} \left[ \sum_m \left[ \trace{K_m} \right]^{\frac{p'}{2}} \right]^{\frac{1}{2p'}}
\end{align}

Since $k_m(x, x') \leq r^2, \ \forall m \in \mathbb{N}_M, \ x \in \mathcal{X}$:

\begin{align}
\label{eq:thm3-10}
& \trace{K_m} \leq Nr^2 \Rightarrow \left[ \trace{K_m} \right]^{\frac{p'}{2}} \leq N^{\frac{p'}{2}}r^{p'} \Rightarrow \nonumber \\
& \Rightarrow \left[ \sum_m \left[ \trace{K_m} \right]^{\frac{p'}{2}} \right]^{\frac{1}{2p'}} \leq M^{\frac{1}{p'}}N^{\frac{1}{4}}r^{\frac{1}{2}}
\end{align}

Thus, combine \eref{eq:thm3-9} and \eref{eq:thm3-10}, we have:

\begin{align}
\label{eq:thm3-11}
\widehat{\Re}_Q(\mathcal{F}) \leq \frac{R}{N}q^{\frac{1}{4}}M^{\frac{1}{2p'}}N^{\frac{1}{4}}r^{\frac{1}{2}} = R\left( \frac{p'}{N^3} \right)^{\frac{1}{4}} \sqrt{rM^{\frac{1}{p'}}}
\end{align}

Combine \eref{eq:thm3-6} and \eref{eq:thm3-11}:

\begin{align}
\label{eq:thm3-12}
& \widehat{\Re}_Q(\boldsymbol{1}^T \bar{\mathcal{F}}) \leq BR\left( \frac{p'}{N^3} \right)^{\frac{1}{4}} \sqrt{rM^{\frac{1}{p'}}} \Rightarrow \nonumber \\
& \Rightarrow \Re_N(\boldsymbol{1}^T\bar{\mathcal{F}}) \leq BR\sqrt{rM^{\frac{1}{p'}} \sqrt{\frac{p'}{N^3}}}
\end{align}

Finally, combine \eref{eq:thm3-5} and \eref{eq:thm3-12}, one can generate the bound provided in \thmref{theorem1}.

\end{proof}

\acresetall

\ifCLASSOPTIONcompsoc
  \section*{Acknowledgments}
\else
  \section*{Acknowledgment}
\fi

Y. Huang acknowledges partial support from a UCF Graduate College Presidential Fellowship and National Science Foundation(NSF) grant No. 1200566. Furthermore, M. Georgiopoulos acknowledges partial support from NSF grants No. 1161228 and No. 0525429, while G. C. Anagnostopoulos acknowledges partial support from NSF grant No. 1263011. Note that any opinions, findings, and conclusions or recommendations expressed in this material are those of the authors and do not necessarily reflect the views of the NSF. Finally, the authors would like to thank the reviewers of this manuscript for their helpful comments.

\bibliographystyle{IEEEtran} 
\bibliography{TPAMI2015paperA}

\end{document}